%% file: main.tex
\documentclass{article}

\usepackage[preprint]{neurips_2026}

\usepackage{amsmath,amssymb}
\usepackage{graphicx}
\usepackage{booktabs}
\usepackage{subcaption}            
\usepackage{xcolor}
\usepackage[hidelinks]{hyperref}   

\newcommand{\vv}{\mathbf{v}}                 
\newcommand{\rL}{\mathbf{r}_{L}}             
\newcommand{\vhat}{\hat{\mathbf{v}}}         
\newcommand{\cosv}{\cos\theta_{v}}           
\newcommand{\Bcoh}{B}                        
\newcommand{\gcoup}{g}                       
\newcommand{\Wdown}{W_{\text{down}}}
\newcommand{\neuron}[2]{\mbox{L#1/F#2}}      
\newcommand{\cstrig}{\cosv^{\mathrm{trig}}}  
\newcommand{\cscoll}{\cosv^{\mathrm{coll}}}  

\title{Leverage Is Not Reach: A Control-Window Law for Single-Neuron Steering in Language Models}

\author{%
  Hongliang Liu\\
  Palo Alto Networks\\
  \texttt{honliu@paloaltonetworks.com}
}

\begin{document}
\maketitle

\input{sections/abstract}
\input{sections/introduction}   
\input{sections/setup}          
\input{sections/law}            
\input{sections/prospective}    
\input{sections/gradient}       
\input{sections/typed_outcomes} 
\input{sections/control_surface}
\input{sections/mechanism}      
\input{sections/unify}          
\input{sections/discussion}     

\bibliographystyle{plainnat}
\bibliography{references}

\input{sections/appendix}       

\end{document}

%% file: sections/abstract.tex
\begin{abstract}
Aligned language models gate behaviors such as refusal, language routing, and arithmetic-operator choice through sparse feed-forward neurons, yet no theory predicts when intervening on a single neuron coherently controls a behavior rather than collapsing the output. Building on forward-only perturbation probing, we develop a budget-normalized control-window framework for single-neuron steering. The effect of one write direction reduces to a single control coordinate, the alignment $\cosv$ between the residual stream and the write, which an injected dose drives along a universal saturation curve in units of a coherence budget $\Bcoh=\lVert\rL\rVert/\lVert\vv\rVert$. Control and collapse are thresholds on this coordinate: coherent control exists exactly when a behavior trigger lies below the collapse ceiling, $\cstrig(\tau)<\cscoll$, the control-window law. The same coordinate governs multiple behaviors, from benign mode switches (language routing, arithmetic-operator choice) to refusal, with a behavior-dependent trigger and the non-safety cases scored by deterministic oracles. The law is predictive but asymmetric: the ceiling scale follows from the weights and one generic forward pass, while the trigger is measured at the behavior's rollout horizon. On fifteen held-out neurons the predicted ceiling has mean absolute error $0.14$ ($\approx0.07$ in bulk layers), and the committed open-or-closed verdict holds on eleven against a ten-of-fifteen majority baseline. A follow-up audit of the closed cases exposes three failure modes rather than violations: the output collapses before the trigger, the write is too late to propagate, or a normalization caps how far one neuron can push. The same law unifies prior single-neuron findings and shows that local gradient attribution \emph{anti-predicts} control, because true controllers write off the readout axis and carry a near-zero first-order gradient; constructively, a forward-only contrastive screen made precise by the window recovers the controllers attribution misses (four of four versus none on matched compute), an efficient recipe for finding behavioral gates. Used as a measurement instrument on refusal, the hardest case, it shows single-neuron intervention success is typed, not scalar: coherent bypass and strict-actionable reach (operationally specific content, not mere non-refusal) separate, so a neuron can flip refusal in fluent, on-task text with no actionable content, and genuine actionable reach appears for only three of six audited Llama pivots and only at later rollout horizons. The framework therefore turns single-neuron steering from a fixed-dose anecdote into a budgeted, typed audit of controllability.
\end{abstract}

%% file: sections/introduction.tex
\section{Introduction}

Alignment fine-tuning installs behaviors such as refusal into language models, and a growing body of work localizes these behaviors to sparse sets of feed-forward (FFN) neurons \citep{geva2021kv,dai2022knowledge}. Causal-circuit analyses localize behaviors to specific attention and FFN components \citep{meng2022rome,wang2023ioi,conmy2023acdc}, and perturbation probing discovers such circuits with two forward passes per prompt \citep{liu2026perturbation}.

Two further lines localize behavior to directions and to single units. Representation-engineering and activation-steering methods steer behavior along residual-stream directions \citep{zou2023repe,rimsky2024caa,arditi2024refusal,subramani2022steering,turner2023actadd,li2023iti}, and single neurons or small circuits have been reported to gate refusal \citep{kazemi2026,herring2026cna,chen2025safety}, to carry out arithmetic \citep{feucht2026}, and to set the residual scale \citep{yu2024superweight}. These results share one striking implication: a single FFN write direction can sometimes flip a model's behavior.

Two problems remain. First, attribution and intervention disagree. The neurons a gradient ranks as important are often not the neurons that causally control a behavior, and neurons that do control it can be invisible to the gradient: a neuron's \emph{leverage}, the ability of a dose or attribution score to move the model state, does not predict its \emph{reach}, the coherent behavioral or semantic effect that actually follows. Second, and more fundamentally, no theory predicts \emph{when} a single-neuron intervention works. The same operation that flips one neuron's behavior coherently drives another into degenerate, repetitive text, and no principle separates the two cases. Both are mechanistic-interpretability problems, since both ask which components \emph{causally} control a behavior rather than merely correlate with it, and the stakes are concrete for safety: a single neuron that flips refusal is a jailbreak vulnerability \citep{kazemi2026,wu2026neurostrike}, and one that cannot is a form of robustness, yet no method says in advance which a given model and neuron will be. This gap has a concrete symptom. Interventions reported at a fixed magnitude give contradictory verdicts, labeling the same neuron controllable or merely destructive depending on a dose chosen by convention. No existing account is at once \emph{causal} (a real intervention, not an attribution score), \emph{dose-aware} (calibrated to the model, not a fixed magnitude), and \emph{predictive} (forecasting controllability before intervening). Without a theory of the dose, single-neuron control remains a catalog of anecdotes rather than a predictable phenomenon, the kind of weights-level, falsifiable regularity a scientific theory of deep learning would require \citep{simon2026theory}. Our picture is a \emph{dose--response} relationship in the pharmacological sense: a single-neuron intervention acts like a compound whose \emph{magnitude}, not merely its presence, sets the outcome --- inert at a low dose, controlling within a therapeutic window, and toxic, collapsing the output, at an overdose. Concretely it is a one-dimensional dial: a dose rotates the model's internal state toward the neuron's write direction, one mark on the dial is where the behavior flips, and another is where the output collapses.

In this work, we present a budget-normalized control-window framework for single-neuron steering. Where activation steering asks which direction changes behavior, we ask when a direction can coherently control it before the model collapses. The effect of a single write direction $\vv$ reduces to one control coordinate, the alignment $\cosv$ between the residual stream and $\vv$. An injected dose drives this coordinate along a universal saturation curve, $\cosv = \gcoup/\sqrt{1+\gcoup^2}$, where $\gcoup = t/\Bcoh$ is the dose in units of a coherence budget $\Bcoh = \lVert\rL\rVert/\lVert\vv\rVert$. Control and collapse are thresholds on this coordinate. Coherent control exists if and only if a behavior-set trigger threshold lies below a collapse ceiling, which we call the \emph{control-window law}. The budget $\Bcoh$ is the FFN-to-skip ratio of \citet{liu2026perturbation} at single-neuron resolution, and the law extends that diagnostic from circuit discovery to a prediction of controllability. Crucially, the ceiling scale follows from the weights and one generic calibration pass, so we forecast the collapse ordering \emph{before} intervening and test the open-or-collapse verdict against a behavior-class trigger estimate, confirming the forecast on held-out neurons across behavior classes. To our knowledge no prior or concurrent method predicts single-neuron controllability. Throughout we study a curated set of controllers chosen to state and test the law; extending the same forward-only primitive to model-scale controller screening across thousands of neurons is the subject of a companion paper.

Our contributions are the following.
\begin{enumerate}
\item \textbf{A control-window law on one coordinate.} The effect of any dose reduces to a single control coordinate, the alignment $\cosv$ between the residual and the write, which follows a universal drive $\cosv=\gcoup/\sqrt{1+\gcoup^2}$ matched to two decimal places across six architectures. Control and collapse are thresholds on this coordinate, and coherent control exists if and only if a behavior-set trigger lies below a weights-predictable collapse ceiling.
\item \textbf{Prospective ceiling prediction and three closed-window mechanisms.} The collapse ceiling, predicted from the weights and a single generic forward pass, is forecast on fifteen held-out neurons with mean absolute error $\mathbf{0.14}$. The closed direction has three mechanisms: a geometric close from a low ceiling, a \emph{drive-capped} close in which a post-FFN normalization caps a single neuron's reach below any trigger (both seen in Gemma-2/3), and a decoupled close at extreme depth.
\item \textbf{Generality across behavior classes.} The same coordinate governs benign operator and language mode switches and refusal, with a behavior-dependent trigger that rises from about $0.3$ for mode switches to $0.45$ for refusal, the mode switches scored by deterministic, judge-free oracles. Concrete gates the recipe finds include \neuron{25}{347} (Qwen3-4B), which routes English output to Chinese; \neuron{19}{13312} (Llama-3.1-8B), which flips $+\!\to\!\times$; the published refusal pivot \neuron{11}{4258} (Llama-3.1-8B; \citealp{kazemi2026}); and the task-framing gate \neuron{8}{3742} (Qwen3.5-2B), which makes the model answer every simple sum with ``mathematically impossible'' ($\cstrig\approx0.6$, a fourth behavior; Appendix~\ref{app:examples}).
\item \textbf{Leverage is not reach.} Local gradient attribution anti-predicts control: true controllers write off the readout axis, pushing sideways in a direction downstream layers amplify rather than one the immediate readout sees, so they carry a near-zero first-order gradient, while gradient-ranked neurons are collapse-prone. The window also has a rollout-time axis; and on refusal, the hardest case, intervention success is typed, not scalar, with coherent bypass failing to predict strict-actionable reach. Constructively, a forward-only contrastive screen made precise by the window recovers the controllers the gradient ranks lowest ($4/4$ vs $0/4$ on matched compute), a backward-pass-free recipe for finding gates.
\item \textbf{A measurement discipline that corrects the record.} Budget-calibrated dosing recovers controllable neurons that fixed-magnitude steering misreports as destructive, and the law places super weights, refusal neurons, and arithmetic and framing gates as points on one window.
\end{enumerate}

Having stated the law, we begin with the object it governs: the single-neuron dose, and the coordinate through which the residual stream reads it.

%% file: sections/setup.tex
\section{Setup: the dose and the control coordinate}
\label{sec:setup}

A single-neuron intervention is one dose along one direction. Neuron $i$ in layer $L$ writes to the residual stream through the column $\vv = \Wdown^{(L)}[:,i]$ of the down-projection, the value side of a feed-forward key-value memory \citep{geva2021kv}. We intervene by adding a scalar multiple of this direction to the residual at layer $L$,
\begin{equation}
\label{eq:ray}
\mathbf{r}(t) = \rL + t\,\vv ,
\end{equation}
where $t$ is the \emph{dose}: the magnitude of the single-neuron intervention. Throughout we use the pharmacological dose--response framing deliberately --- the \emph{magnitude}, not the mere presence, of the intervention determines whether the effect is inert, controlling, or toxic (a collapse) --- which is why a dose calibrated to the model, rather than a fixed magnitude, is the right unit. Setting the neuron's activation to a value (pinning) and adding to its output (injection) move the state along the same ray~\eqref{eq:ray}, so we treat them as one operation and report results in terms of $t$. This equivalence holds for the pre-norm architectures we study; under a post-FFN normalization (Gemma-2/3) the neuron's output is renormalized before it joins the residual, capping how far pinning can move the state and breaking the equivalence, which we isolate as a third closed-window mechanism (Section~\ref{sec:prospective}, Appendix~\ref{app:geom}).

\begin{figure}[t]
\centering
\includegraphics[width=0.6\linewidth]{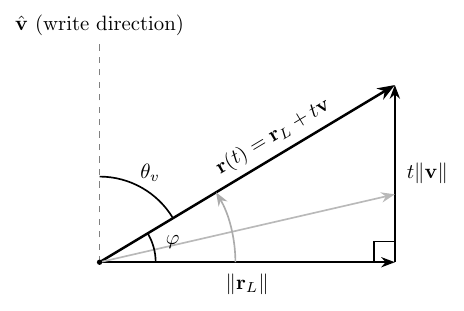}
\caption{The geometry of a single-neuron dose. Because the write direction $\vhat$ is orthogonal to the residual $\rL$, the dose $t\vv$ builds a right triangle whose hypotenuse is the perturbed residual $\mathbf r(t)=\rL+t\vv$ (Equation~\ref{eq:ray}). Increasing the dose rotates $\mathbf r(t)$ from $\rL$ toward $\vhat$. The control coordinate $\cosv$ (Equation~\ref{eq:cosv}) is the cosine of the angle $\theta_v$ between $\mathbf r(t)$ and $\vhat$, and the coupling $\gcoup=t/\Bcoh$ (Equation~\ref{eq:g}) is the tangent of the rotation angle $\varphi$ away from $\rL$. The dose $t=\Bcoh$ is the $45^\circ$ point where the write matches the residual.}
\label{fig:geometry}
\end{figure}

The residual stream enters every downstream computation through a normalization \citep{ba2016layernorm,zhang2019rmsnorm}, which discards magnitude and keeps direction. The dose both rotates $\mathbf{r}(t)$ toward $\vhat$ and inflates its norm, but the normalization discards the inflation, so only the rotation survives; this is why the readout reduces to a single coordinate. That coordinate is the direction of $\mathbf{r}(t)$, summarized by one scalar, the alignment of that direction with the write direction,
\begin{equation}
\label{eq:cosv}
\cosv \;=\; \cos\angle\!\big(\mathbf{r}(t),\,\vhat\big), \qquad \vhat = \vv/\lVert\vv\rVert .
\end{equation}
We call $\cosv$ the \emph{control coordinate}. It is signed along the dosing ray; where control is sign-specific we report $|\cosv|$ at the behavior boundary and label the controlling branch ($+\vv$ or $-\vv$), and for the few neurons whose baseline is not orthogonal (e.g.\ \neuron{11}{4258}, $\cosv(0)=0.36$) the boundary is measured from that baseline on its controlling branch. Because the write direction is, at baseline, nearly orthogonal to the residual it is added to (Section~\ref{sec:law}), the dose enters $\cosv$ only through the dimensionless ratio
\begin{equation}
\label{eq:g}
\gcoup \;=\; \frac{t\,\lVert\vv\rVert}{\lVert\rL\rVert} \;=\; \frac{t}{\Bcoh}, \qquad \Bcoh = \frac{\lVert\rL\rVert}{\lVert\vv\rVert} .
\end{equation}
Here $\gcoup$ is the magnitude the neuron writes into the residual relative to the residual it writes into, which is the FFN-to-skip ratio of \citet{liu2026perturbation} evaluated for a single neuron, and $\Bcoh$ is the dose at which that ratio reaches one. We call $\Bcoh$ the \emph{coherence budget}; $\lVert\rL\rVert$ is measured once on generic text at generation positions. The three quantities are the sides and angles of one right triangle: because the write direction is orthogonal to the residual, Equations~\eqref{eq:ray}--\eqref{eq:g} are the geometry of Figure~\ref{fig:geometry}, in which the dose rotates the residual toward $\vhat$.

We obtain candidate controllers by contrastive discovery, the forward-only primitive of \citet{liu2026perturbation}: for a target behavior we run a small set of contrasting prompts and rank neurons by the difference in their activations, then verify causality by dosing. This requires only forward passes and reuses published anchors where available \citep{yu2024superweight,kazemi2026,feucht2026}. Unless attributed to one of these anchors, the neurons we report are identified in this work; we name each by its layer and index together with its model, for example the operator gate \neuron{19}{13312} on Llama-3.1-8B. The behaviors we study are categorical or template-like: refusal, language routing (English to Chinese), arithmetic-operator choice (addition to multiplication), and a task-framing exemplar (locking the response into a fixed template). Throughout, every behavior flip is read under a coherence gate matched to the output script, using the distinct-bigram ratio for whitespace-delimited generations and character diversity for unspaced scripts such as Chinese, and at a dose calibrated in units of $\Bcoh$ rather than at a fixed magnitude; Section~\ref{sec:contract} details this measurement and the reporting errors it corrects.

Table~\ref{tab:notation} collects the notation used throughout. Having defined the dose and the coordinate it drives, we now state the law that governs them.

\input{sections/notation}

%% file: sections/notation.tex
\begin{table}[t]
\centering\small
\caption{Notation. The control window is one inequality on one coordinate: coherent control exists when the
behavior trigger lies below the collapse ceiling, $\cstrig(\tau)<\cscoll$. Geometry sets $\cosv$; the budget and
collapse coefficient set its scale; the trigger is behavior-set.}
\label{tab:notation}
\begin{tabular}{@{}ll@{}}
\toprule
symbol & meaning \\
\midrule
\multicolumn{2}{@{}l}{\emph{intervention and coordinate}}\\
$t$            & \emph{dose}: magnitude of the single-neuron intervention (the steering scalar we vary) \\
$\vv,\ \vhat$  & neuron's write direction ($\Wdown$ column) and its unit vector \\
$\rL$          & residual stream at the intervention layer $L$ \\
$\cosv$        & \emph{control coordinate}: $\cos\angle(\rL+t\vv,\ \vhat)$, the one quantity the normalized readout sees \\
\midrule
\multicolumn{2}{@{}l}{\emph{scales (from the weights)}}\\
$\Bcoh$        & \emph{coherence budget} $=\lVert\rL\rVert/\lVert\vv\rVert$: dose at which the write rivals the residual (FFN-to-skip ratio) \\
$\gcoup$       & \emph{coupling} $=t/\Bcoh$: dimensionless dose; drives $\cosv=\gcoup/\sqrt{1+\gcoup^2}$ \\
$m^\ast$       & \emph{collapse coefficient} $=t^\ast/\Bcoh$: order-one, architecture-family-dependent; sets the ceiling \\
$\mathrm{PR}$  & participation ratio of $\rL$: residual effective dimensionality; sets $m^\ast$ \\
\midrule
\multicolumn{2}{@{}l}{\emph{window edges}}\\
$\cstrig(\tau)$ & \emph{trigger}: $\cosv$ at which the behavior flips (behavior- and rollout-dependent) \\
$\cscoll$       & \emph{collapse ceiling}: $\cosv$ at which the output degenerates ($=m^\ast/\sqrt{1+m^{\ast2}}$) \\
\midrule
\multicolumn{2}{@{}l}{\emph{rollout, attribution, composition}}\\
$\tau,\ \tau^\ast$ & rollout horizon / onset time (tokens): where the behavior is read / first becomes visible \\
$\alpha,\ \beta$   & first- (gradient) and second-order (curvature) coefficients of the behavior margin along $\vv$ \\
$\lvert\Delta a\rvert$ & contrastive activation change used to screen candidate neurons \\
$K$            & number of simultaneously dosed neurons ($K{=}1$: single-neuron) \\
\bottomrule
\end{tabular}
\end{table}

%% file: sections/law.tex
\section{The control-window law}
\label{sec:law}

Coherent single-neuron control exists if and only if the dose that flips the behavior is smaller than the dose that collapses the output. On the control coordinate, this is a window: a behavior-set \emph{trigger} $\cstrig(\tau)$, which depends on the rollout length $\tau$ over which the behavior is read, and a coherence-set \emph{collapse} ceiling $\cscoll$, with
\begin{equation}
\label{eq:law}
\text{coherent control} \iff \cstrig(\tau) < \cscoll .
\end{equation}
We call Equation~\eqref{eq:law} the control-window law. In words: the ceiling is a geometric budget, the trigger is a behavior-set boundary, and coherent control exists when, and only when, the trigger falls below the ceiling; Section~\ref{sec:prospective} tests both directions, with open windows yielding coherent control and closed windows, from a low ceiling or an unreachable trigger, yielding none. The coordinate is signed, and the law is read \emph{branch-resolved}: the trigger and the ceiling entering Equation~\eqref{eq:law} are measured on the same controlling sign of $\vv$, never compared across branches, because the non-controlling branch can collapse without any behavior flip and a sign-blind ceiling search then reports that collapse as a spuriously low ceiling (Section~\ref{sec:contract}, Appendix~\ref{app:prosp}). The trigger is a rollout-time boundary, not a scalar: the single number we quote for a behavior is the value of $\cstrig(\tau)$ at the verification horizon used for that behavior, and Section~\ref{sec:surface} measures the full $\cstrig(\tau)$ as the dose-by-rollout-time surface of which this window is the dose projection. The window width $\cscoll - \cstrig(\tau)$ is the controllability margin. The rest of this section defines the three quantities the law is built from: the universal drive that maps dose to coordinate, the ceiling, and the trigger.

As written, the law is close to a definition: coherent control means the behavior flips while the output stays coherent, which is exactly the coordinate lying between the trigger and the ceiling. Its empirical content is therefore not the inequality but the claim that both edges are fixed in advance. The ceiling scale follows from the weights and a generic forward pass (Section~\ref{sec:prospective}) and the trigger from behavior class, depth, and rollout horizon, so the window can be computed before any intervention and the open-or-closed verdict tested prospectively. Without that independent predictability the law would be a tautology; with it, each held-out neuron is a falsifiable prediction, which is how we test it in Section~\ref{sec:prospective}. The two edges are not equally constrained: the ceiling we predict from geometry, to a parameter-free scale times an order-one family coefficient, whereas the trigger we predict only from behavior class and rollout horizon, not from the weights. The theory thus constrains collapse more tightly than control; Section~\ref{sec:mechanism} takes a first step toward a trigger theory, and a fully first-principles trigger value remains open. We fix the falsifier in advance: the law is falsified by a neuron whose trigger is measured below its predicted ceiling on the controlling branch under a budgeted dose ladder yet shows no coherent control, or whose output collapses far below its family- and depth-calibrated ceiling. A neuron that shows no behavioral response at \emph{any} coherent dose is recorded, by this pre-registered criterion, as a \emph{decoupled close} rather than an open-window failure, because no trigger is reachable to test the inequality; this is a distinct outcome, not a post-hoc rescue. Across the fifteen held-out neurons none met the falsifier.

\subsection{The drive is a universal saturation curve}
The control coordinate is a deterministic function of the coupling. Since the write direction is orthogonal to the residual, the perturbed residual is the hypotenuse of a right triangle with legs $\lVert\rL\rVert$ and $t\lVert\vv\rVert$ (Figure~\ref{fig:geometry}), so
\begin{equation}
\label{eq:drive}
\cosv = \frac{\gcoup}{\sqrt{1+\gcoup^2}} .
\end{equation}
Equation~\eqref{eq:drive} is a mixing angle: the dose rotates the residual from its input direction toward $\vhat$, and $\gcoup$ is the tangent of that rotation. In these terms control is a rotation of roughly $17$ to $56^\circ$ of the state toward $\vhat$ ($\cosv=\sin\varphi$ with $\varphi=\arctan\gcoup$), the budget $\Bcoh$ being the $45^\circ$ equal-mix dose, while Gemma collapses near $17^\circ$, the rotation other families need merely to reach the trigger. The drive is universal. Measured across six architectures and a dose sweep, $\cosv$ matched Equation~\eqref{eq:drive} to two decimal places at every dose, with $\cosv\approx0.00$ at $t=0$, confirming both the orthogonality assumption and that the model adds the write linearly within the forward pass. A contrastively active controller is not exempt: its own baseline contribution along $\vhat$ is of order $a_i/\Bcoh$ (the neuron's baseline activation in units of the budget), small whenever the budget exceeds that activation, so even selected controllers sit near $\cosv(0)\approx0$; the exceptions are neurons with a large baseline activation relative to their budget, such as the small-budget \neuron{11}{4258} ($\cosv(0)=0.36$) and \neuron{20}{1378} ($\cosv(0)=0.24$).\footnote{For a baseline alignment $c_0=\cosv(0)\neq0$ the exact signed drive is $\cosv(g) = (c_0+g)/\sqrt{1+2c_0 g+g^2}$, which reduces to Equation~\eqref{eq:drive} at $c_0=0$. We use the simple curve where $c_0\approx0$, which is the measured case for all but a few neurons, and the offset-corrected branch coordinate for the high-baseline exceptions.} The expansion $\cosv = \gcoup - \tfrac12\gcoup^3 + \cdots$ converges for $\gcoup<1$, so the perturbative regime is exactly $t<\Bcoh$. This universality is a property of the coordinate, measured by injecting the write into the residual; whether a single neuron can drive the coordinate by its own activation is architecture-dependent, and under a post-FFN normalization (Gemma-2/3) it is capped below any trigger, a third way the window closes that we take up in Section~\ref{sec:prospective} and Appendix~\ref{app:geom}.

\subsection{The ceiling is the coherence budget}
The output collapses when the dose reaches a fixed multiple of the coherence budget. Beyond that dose the normalized residual is pinned to $\vhat$ regardless of input, and the generation degenerates into repetition or scramble. Writing the collapse dose as $t^\ast = m^\ast \Bcoh$, the collapse ceiling is $\cscoll = m^\ast/\sqrt{1+m^{\ast 2}}$. The budget $\Bcoh$ supplies the parameter-free \emph{scale} of collapse; the coefficient $m^\ast$, the collapse magnitude in units of $\lVert\rL\rVert$, is an order-one number set per architecture family by how concentrated the residual is (its participation ratio), with a weaker dependence on depth. It is a small calibrated correction, not a free per-neuron fit: $m^\ast$ is $1.4$--$1.7$ across the Qwen and Llama families and $0.31$ for the near-rank-one Gemma residual. We therefore predict the collapse \emph{ordering} of held-out neurons from the weights, validated prospectively in Section~\ref{sec:prospective}, rather than the absolute collapse dose to high precision. The budget grows with depth, so controllability concentrates in late layers, where a neuron has room to flip a behavior before collapse.

\subsection{The trigger is behavior-dependent}
The behavior flips when the rotation crosses a basin boundary, at a coordinate $\cstrig(\tau)$ set by the behavior rather than the geometry. The trigger is behavior- and readout-dependent and has two regimes (Section~\ref{sec:mechanism}, Appendix~\ref{app:trigger}): for token-local routers it can be read as a branch-resolved first-token margin crossing, while for multi-token behaviors such as operator switching it develops over a short autoregressive rollout. We treat its value as measured (Section~\ref{sec:surface}), not predicted from a closed-form curvature law. Its value and sharpness depend on the behavior class. Read at a short, token-local rollout, categorical mode switches such as language routing and operator choice flip sharply near $\cstrig\approx0.3$, while coarse decisions such as refusal ramp softly and partially, reaching half their effect near $\cstrig\approx0.45$ (measured in Section~\ref{sec:prospective}), and task-framing controllers that lock the response into a fixed template trigger higher still, near $\cstrig\approx0.6$ (Appendix~\ref{app:examples}). These are values of $\cstrig(\tau)$ at one horizon; Section~\ref{sec:surface} shows the boundary moves with $\tau$, declining toward a lower asymptote for a token-local router and sitting on a hard rollout floor for the operator. This extends the opposition-versus-routing circuit distinction of \citet{liu2026perturbation} from a structural label to a quantitative, rollout-resolved threshold.

\begin{figure}[t]
\centering
\includegraphics[width=0.62\linewidth]{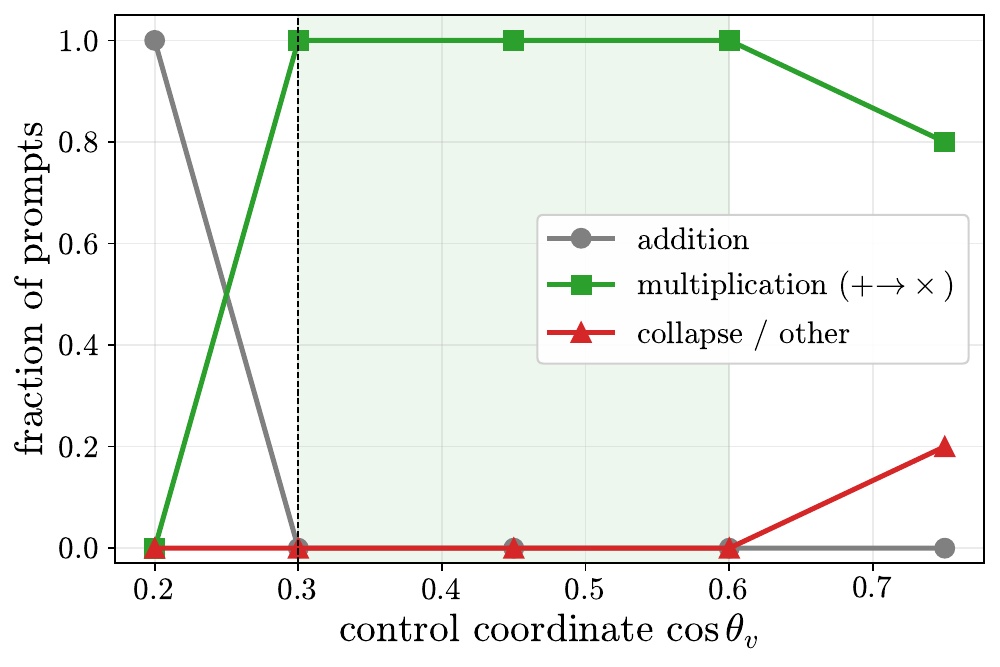}
\caption{The control window on a benign, judge-free behavior. An operator neuron (Llama \neuron{19}{13312}) is dosed on symbolic arithmetic $a+b{=}$ and scored by exact match: addition (gray) holds below the trigger, the neuron switches addition to multiplication ($+\!\to\!\times$, green) sharply from $\cosv\approx0.3$, and the output begins to collapse at the largest doses shown ($\cosv\approx0.75$), approaching the ceiling near $0.84$ (red). The exact oracle needs no coherence gate, isolating the window from the measurement corrections that safety behaviors require (Section~\ref{sec:contract}).}
\label{fig:operator}
\end{figure}

Figure~\ref{fig:operator} shows the window in its simplest measurable form, on a benign behavior whose oracle is exact. Injecting an operator neuron on symbolic arithmetic, the model holds addition below the trigger, switches cleanly to multiplication across the window from $\cosv\approx0.3$, and collapses past the ceiling. This is an operator-\emph{selector} gate that decides $+\!\to\!\times$, distinct from and complementary to the prior base-10 addition \emph{mechanism} of \citet{feucht2026}. Because the behavior is scored by exact match rather than a coherence judge, the window is visible with no measurement correction at all, which makes the operator the cleanest instance of the law and shows it is not specific to safety. It is also a controller surfaced by a two-forward-pass screen and confirmed by the window, with no gradient or activation patching. A higher-trigger task-framing exemplar (Qwen \neuron{8}{3742}, which locks the model into declaring a simple sum ``mathematically impossible'') obeys the same law judge-free by template uniformity at a higher trigger (Appendix~\ref{app:examples}). Across benign mode switches and refusal, the law is measured behavior by behavior, with refusal the stress test rather than the subject.

\subsection{Measurement contract for a control window}
\label{sec:contract}
Because the law compares two edges of a signed window, every control measurement follows five conventions, summarized in Table~\ref{tab:contract}: a budgeted dose in units of $\Bcoh$, a signed controlling branch, the behavior's own rollout horizon $\cstrig(\tau)$, a script-matched coherence gate, and, for safety, a content gate that reports strict-actionable content separately from non-refusal. These are not extra assumptions but the conditions under which the window's two edges are measured on the same object; the artifacts that motivate each, and the judge details, are in Appendix~\ref{app:measure}.

\begin{table}[t]
\centering
\small
\caption{The measurement contract: the five conventions under which a control window is measured. Artifact analysis and judge details are in Appendix~\ref{app:measure}.}
\label{tab:contract}
\begin{tabular}{p{0.26\linewidth} p{0.64\linewidth}}
\toprule
convention & what it fixes \\
\midrule
Budgeted dose & dose swept in units of $\Bcoh$, not a fixed magnitude; comparable across neurons, avoids fixed-dose overshoot \\
Signed branch & trigger and ceiling on the same controlling sign of $\vv$; a non-controlling-branch collapse is not the ceiling \\
Rollout horizon & behavior read at its own horizon, so the trigger is $\cstrig(\tau)$, not a scalar \\
Coherence gate & a flip counts only while the output is coherent, via a script-matched diversity gate \\
Content gate (safety) & non-refusal is not harm; strict-actionable $=$ coherent $\wedge$ on-task $\wedge$ complies $\wedge$ actionable, reported separately from non-refusal \\
\bottomrule
\end{tabular}
\end{table}

Equations~\eqref{eq:law}--\eqref{eq:drive} reduce controllability to comparing two thresholds on one universal curve. We treat the window here as an interval on the dose coordinate; Section~\ref{sec:surface} shows that the full object is a dose-by-rollout-time surface, of which this interval is the dose projection. We next show that one of the two thresholds, the ceiling scale, is predictable from the weights and a generic forward pass, and test the law prospectively.

%% file: sections/prospective.tex
\section{The control window is predictable: prospective validation}
\label{sec:prospective}

The control window can be computed before any intervention, from the weights and a single generic forward pass: its ceiling from residual geometry, and its open-or-closed verdict from the law. We validate this in two stages: a held-out test that the budget sets the collapse dose, and a predict-then-dose test of the full window.

\subsection{Collapse is universal in the rescaled coupling}
\begin{figure}[t]
\centering
\includegraphics[width=\linewidth]{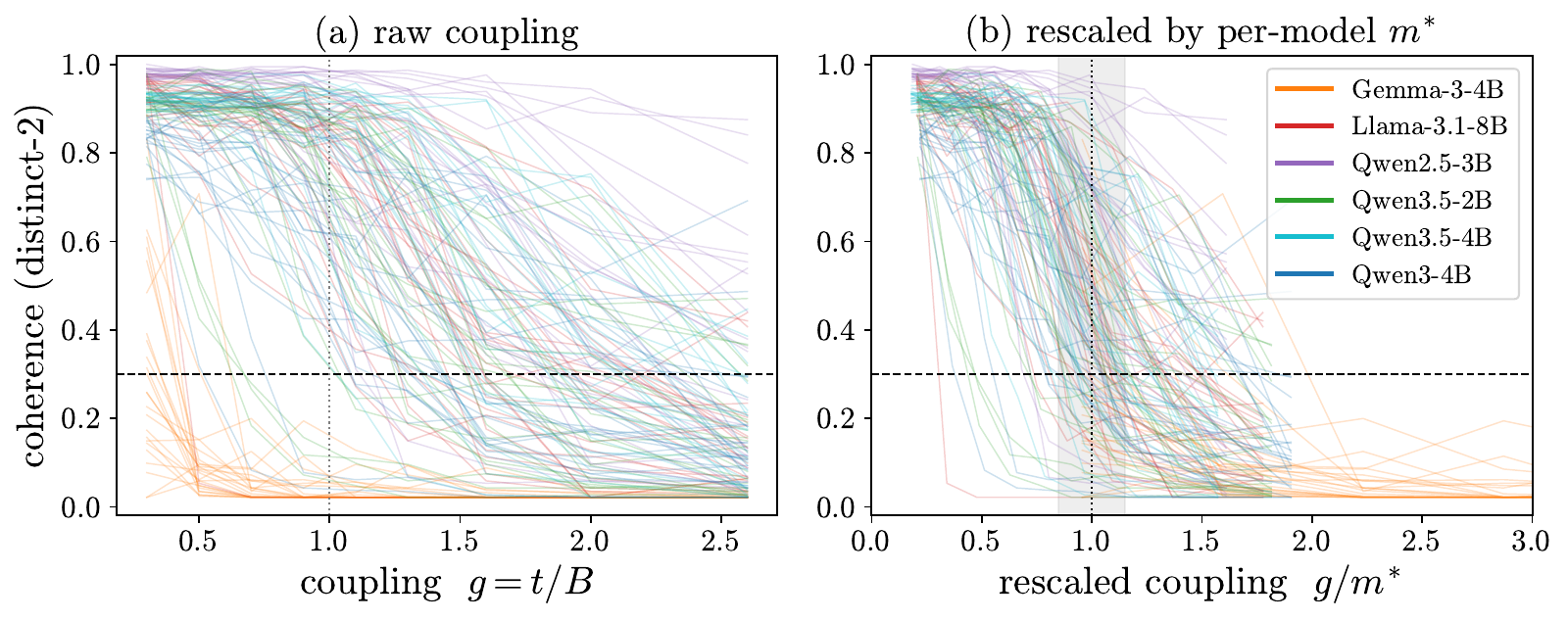}
\caption{Collapse is universal in the rescaled coupling. (a) Coherence, the distinct-bigram ratio, against the coupling $\gcoup=t/\Bcoh$ for the held-out neurons across six architectures; the collapse transition is scattered, near $\gcoup\approx1.5$ for the Qwen and Llama families and $\approx0.3$ for Gemma. (b) Rescaling the coupling by each model's single constant $m^\ast$ collapses every curve onto one cliff at $x=1$ (shaded band). The dashed line is the $0.30$ coherence floor.}
\label{fig:collapse}
\end{figure}

On neurons never used to derive it, collapse occurs at a fixed multiple of the budget. We drew $24$ random neurons per model across six architectures, predicted each collapse dose as $\Bcoh$ from weights and a single generic forward pass, then dosed and located collapse by the distinct-bigram ratio. The collapse coefficient $m^\ast = t^\ast/\Bcoh$ fell in a tight band of $1.37$ to $1.65$ across the five Qwen and Llama models (within-model $R^2$ from $0.40$ to $0.97$), with Gemma-3 the lone outlier at $m^\ast=0.31$. Rescaling each curve by its model's mean $m^\ast$ collapsed all curves, across all six architectures, onto a single near-vertical cliff (Figure~\ref{fig:collapse}). The transition is sharp because collapse is a saturation: once the injected direction dominates, it dominates abruptly.

\begin{figure}[t]
\centering
\includegraphics[width=0.66\linewidth]{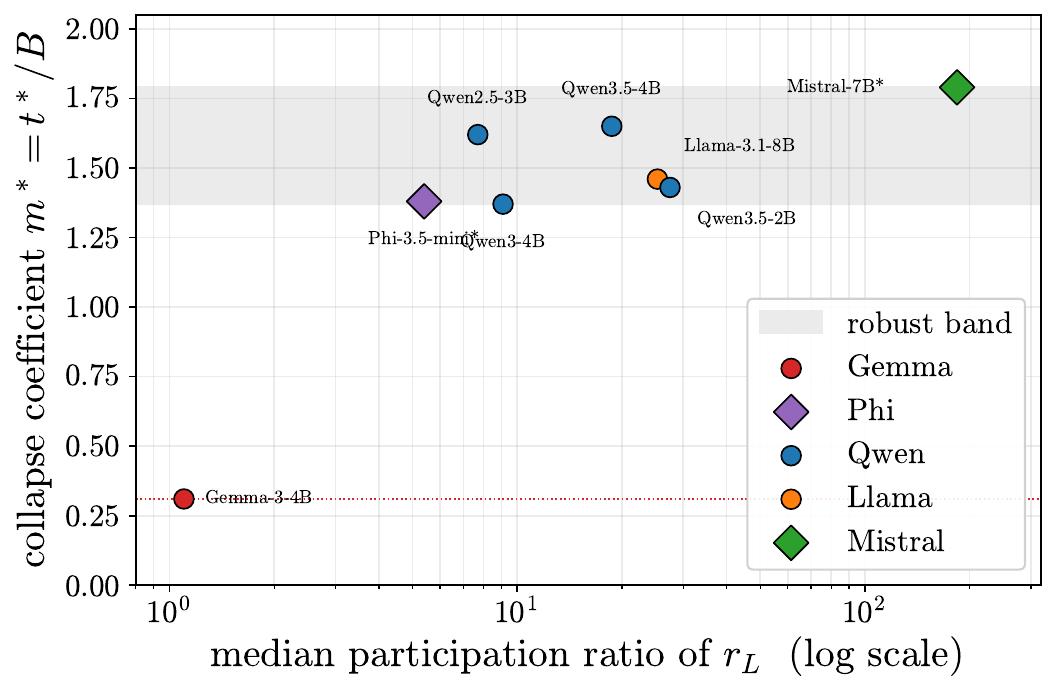}
\caption{The collapse coefficient tracks residual concentration across eight architectures (Spearman $0.67$). The seven distributed-residual models occupy a robust band ($m^\ast\approx1.4$ to $1.8$); only Gemma's near-rank-one residual makes it fragile. Phi (held out, diamond) has a concentrated, massive-activation residual (PR $5.4$) yet is robust, so the fragility is the near-rank-one limit, not low PR in general. Stars in Table~\ref{tab:mstar} mark the held-out families.}
\label{fig:prmstar}
\end{figure}

Two honest corrections follow from the held-out data. The coefficient is $\approx 1.5$, not one, and it is not universal: Gemma-3 collapses at $m^\ast=0.31$, five times more fragile. A pooled cross-model regression returns $R^2=0.98$, but this is an artifact of Gemma's much larger budget anchoring the fit, and we discard it. The mechanism of the family dependence is concentration. Gemma's residual has a median participation ratio of $1.1$, meaning its norm is carried by essentially one dimension, so a generic-direction write of a small fraction of $\lVert\rL\rVert$ already overwhelms the computation-bearing subspace. The collapse coefficient therefore tracks the participation ratio of $\rL$, which is itself a weights-and-one-pass quantity. Two further held-out families confirm and sharpen this. Across all eight architectures $m^\ast$ rises with PR (Spearman $0.67$, not significant at $n=8$; Table~\ref{tab:mstar}, Figure~\ref{fig:prmstar}), but the relation is a near-rank-one cliff, not a graded predictor: Mistral, whose residual is highly distributed (PR $184$), sits at the top of the robust band ($m^\ast=1.79$), and Phi-3.5 is the informative case, with a concentrated residual carrying massive activations (PR $5.4$) yet still robust ($m^\ast=1.38$). PR thus separates the fragile rank-one regime, reached only by Gemma, from a robust band within which $m^\ast$ still needs a per-family and weak per-depth calibration; it is not low PR in general that predicts fragility. Phi's single-constant fit is correspondingly the loosest ($R^2=0.25$), reflecting the depth dependence of $m^\ast$ that the bulk-depth restriction of the predict-then-dose test below removes.

\begin{table}[t]
\centering
\caption{Collapse coefficient $m^\ast=t^\ast/\Bcoh$ on held-out neurons, by architecture, with the median participation ratio (PR) of the residual. Across eight architectures $m^\ast$ tracks PR (Spearman $0.67$): the seven distributed-residual models occupy a robust band, and Gemma's near-rank-one residual makes it five times more fragile. Mistral and Phi are held-out families; Phi shows that a concentrated residual (PR $5.4$) is still robust, so the fragility is the near-rank-one limit, not low PR in general.}
\label{tab:mstar}
\begin{tabular}{lccc}
\toprule
model & median PR & mean $m^\ast$ & within-model $R^2$ \\
\midrule
Qwen3-4B & 9.1 & 1.37 & 0.76 \\
Qwen3.5-2B & 27.5 & 1.43 & 0.48 \\
Qwen3.5-4B & 18.7 & 1.65 & 0.71 \\
Qwen2.5-3B & 7.7 & 1.62 & 0.50 \\
Llama-3.1-8B & 25.3 & 1.46 & 0.40 \\
Mistral-7B & 184 & 1.79 & 0.74 \\
Phi-3.5-mini & 5.4 & 1.38 & 0.25 \\
Gemma-3-4B & 1.1 & \textbf{0.31} & 0.97 \\
\bottomrule
\end{tabular}
\end{table}

\subsection{Predict-then-dose: the control window}
The ceiling can be fixed ahead of dosing on held-out neurons, and the open-or-closed verdict prospectively tested against a behavior-class trigger estimate. We committed every such prediction before dosing, the ceiling from geometry and the trigger from behavior class, then dosed fifteen held-out safety controllers across five architectures and scored the verdicts blind. No held-out neuron met the falsifier after the branch-resolved audit, and the committed verdict was correct on eleven; because the held-out set is ten-open and five-closed, this only marginally beats a ten-of-fifteen majority-class baseline, so we treat the continuous ceiling error, not the verdict count, as the prospective result. The collapse ceiling was predicted with a mean absolute error of $0.14$ (bootstrap $95\%$ confidence interval $0.09$ to $0.23$); within the network bulk, at relative depth below $0.8$ where controllers live, the error fell to about $0.07$. One committed open neuron carried a nominal trigger-above-ceiling inversion that a branch-resolved re-measurement resolved to a clean open, tracing it to an output-diversity collapse on the non-controlling branch and placing the controlling-branch ceiling above the trigger (Appendix~\ref{app:prosp}). The four verdict misses were not law violations but informative edges, which we take in turn.

The closed direction has three mechanisms. On Gemma, predicted closed and confirmed, the collapse ceiling is the lowest we measure, $\cscoll\approx0.30$ ($m^\ast=0.31$), below any safety trigger: the \emph{geometric} closed window, $\cscoll<\cstrig$. That ceiling, however, is reached only by direct injection; a single neuron never gets there, because Gemma's post-FFN normalization caps its own activation's reach to $\cosv\le0.19$ at any pin, with coherence intact (Appendix~\ref{app:geom}). This \emph{drive-capped} close is the operative one for single-neuron control: the neuron cannot reach its coordinate even though the coordinate exists. It scopes the pinning--injection equivalence of Section~\ref{sec:setup} to pre-norm models (the architectural-defense reading is in Appendix~\ref{app:geom}). In the final layers of the Qwen networks, above relative depth $0.9$, two neurons predicted open came out closed for a third reason: the behavior never responded to any coherent dose, because too few layers remain after the write to propagate it, so there is effectively no trigger; the \emph{decoupled} closed window. The three mechanisms, geometric, drive-capped, and decoupled, give the closed direction a structure the single geometric case it previously rested on did not capture.

A dosing artifact was recovered. Llama \neuron{31}{3309}, predicted an artifact because fixed-magnitude pinning showed mode collapse, was instead open: the calibrated dose sweep revealed a clean window on $-\vv$. This is a second prospective instance of the dose-calibration correction, and it cost us a committed verdict precisely because the effect is real.

The trigger is also U-shaped in depth: lowest near the network middle ($\approx0.4$) and rising toward both ends (the early-layer \neuron{8}{397} at $0.67$), so it depends on depth as well as behavior class. The remaining miss, Llama \neuron{24}{2598}, was a weak open whose peak bypass of $0.25$ fell just under the coherence gate, a threshold call rather than a law failure. Table~\ref{tab:prospective} summarizes; the full fifteen-neuron table is Table~\ref{tab:r1full} in the appendix.

\begin{table}[t]
\centering
\caption{Prospective control-window test, summary (per-neuron predictions and dosed outcomes for all fifteen are in Table~\ref{tab:r1full}). Every prediction was committed from weights and behavior class before dosing, then scored blind. The continuous ceiling error, not the verdict count, is the prospective result.}
\label{tab:prospective}
\begin{tabular}{@{}l p{0.5\linewidth}@{}}
\toprule
quantity & value \\
\midrule
held-out neurons (open / closed) & $15$ ($10$ / $5$) \\
committed open-or-closed verdict correct & $\mathbf{11/15}$ \\
majority-class baseline & $10/15$ \\
ceiling MAE, all layers & $\mathbf{0.14}$ ($95\%$ CI $0.09$--$0.23$) \\
ceiling MAE, bulk (rel.\ depth $<0.8$) & $\approx0.07$ \\
verdict misses (all law-consistent) & $4$: geometric and decoupled closes, a recovered dosing artifact, and a sub-gate weak open \\
\bottomrule
\end{tabular}
\end{table}

Having shown the law is predictive across fifteen held-out neurons, we turn to its sharpest consequence, that the gradient predicts the wrong quantity.

%% file: sections/gradient.tex
\section{Discovery: salience finds candidates, the law finds controllers}
\label{sec:gradient}

Local gradient attribution misses controllers for two separable reasons. The first is geometric: the per-dose rotation rate of the control coordinate at the origin is
\begin{equation}
\label{eq:slope}
\left.\frac{d\cosv}{dt}\right|_{t=0} = \frac{1}{\Bcoh} = \frac{\lVert\vv\rVert}{\lVert\rL\rVert},
\end{equation}
so $1/\Bcoh$ is the rate at which the dose drives the coordinate toward collapse, and a large-norm write spends the coherence budget fastest. A gradient-attribution score weights a neuron by the behavior gradient $\alpha=\nabla_{\rL}M\!\cdot\!\vv$ (with $M$ the behavior margin), which scales with $\lVert\vv\rVert$; within a layer (common $\lVert\rL\rVert$) the neurons it ranks highest are therefore the high-$\lVert\vv\rVert$, low-$\Bcoh$, collapse-prone ones. We distinguish this geometric leverage, $1/\Bcoh$, from the behavior-gradient score $\alpha$ it correlates with through $\lVert\vv\rVert$; the two coincide only within a layer. The second reason is the load-bearing one, the \emph{silent gate}: a true controller writes \emph{off} the behavior readout axis (Section~\ref{sec:mechanism}), so its first-order behavior gradient $\alpha\approx0$ and attribution sees nothing, while the flip is carried by the finite-dose curvature $\beta=\vv^\top H\vv$ that the linear term cannot access. The two reasons are independent: the first puts collapse-prone neurons at the top of the gradient ranking, the second hides the genuine controllers at the bottom. Leverage and reach are inverse on the control coordinate for both.

This is visible directly. The highest-gradient neurons collapse the model, and the clean controllers carry the smallest gradients. A taxonomy on the first- and second-order channels separates neurons by \emph{readability} (whether attribution sees them), not by causal use \citep{geiger2021causal}, so it functions as a causal filter rather than a ranker. A filter that removes high-leverage (high-$1/\Bcoh$) neurons removes the collapse-prone ones and retains the controllers, which is why it outperforms a gradient ranker on the same task.

A same-task comparison makes the cost concrete. Table~\ref{tab:h2h} reports a head-to-head on Qwen2.5-3B, scoring the top neurons of a gradient ranker against the top neurons surviving the causal filter, both validated by generation under the register-aware judge (Section~\ref{sec:contract}).

\begin{table}[t]
\centering
\caption{Gradient ranking versus the causal filter on the same task and compute (Qwen2.5-3B, coherence-gated bypass judge: coherent $\wedge$ non-refusal, Claude-Haiku; Section~\ref{sec:typed}). Counts are coherent non-refusal (behavior-level bypass), not genuine-harm compliance (Section~\ref{sec:unify}), out of 24 prompts.}
\label{tab:h2h}
\begin{tabular}{lcc}
\toprule
metric & gradient top-20 & filter top-19 \\
\midrule
known pivots recovered & 0/4 & \textbf{4/4} \\
best single-neuron bypass & 6/24 & \textbf{20/24} \\
successful neuron-prompt pairs & $18/480$ & $\mathbf{54/456}$ \\
inert (Class C/D) in selection & 15/20 & \textbf{0/19} \\
\bottomrule
\end{tabular}
\end{table}

The gradient top-20 recovered none of the four known controllers and selected inert neurons three quarters of the time; the filter recovered all four and surfaced the strongest single-neuron behavior bypass measured in this work, \neuron{21}{2722} at $20/24$ (coherence-gated bypass judge). This bypass is coherent non-refusal, and under content audit \neuron{21}{2722} produces no genuine harm on either sign (Section~\ref{sec:unify}); it is thus simultaneously the strongest controller the gradient misses and the clearest case that control is not harm. The most controllable neurons are invisible to the gradient by construction. A language router we identify (\neuron{25}{347}, Qwen3-4B) sits in the bottom few percent of the behavior-token gradient yet switches language on $12/12$ prompts. A silent-gate operator neuron (\neuron{19}{13312}, Llama-3.1-8B) is missed by both gradient and subspace attribution yet flips addition to multiplication; in a controlled $61$-neuron pool the local gradient ranks it last ($\alpha=-0.08$, noise) while the curvature $\beta$ ranks it first, which explains the blindness. Curvature is not a discovery ranker either: a $200$-neuron operator-$\beta$ scan nominates high-curvature Qwen neurons that remain causally inert at $K{=}1$ (Appendix~\ref{app:trigger}), so $\beta$ explains a known silent controller but, ranked alone, surfaces a causally inert sector. These are not edge cases but the rule implied by Equation~\eqref{eq:slope}: a discovery method that ranks leverage ranks collapse-proneness, and the controllers live where leverage is low. We establish this for the local next-token gradient leverage under the budgeted protocol tested here; we do not claim it for rollout-differentiated or integrated-gradient rankers, which a fuller baseline comparison would weigh.

\begin{figure}[t]
\centering
\begin{subfigure}{0.49\linewidth}
  \centering
  \includegraphics[width=\linewidth]{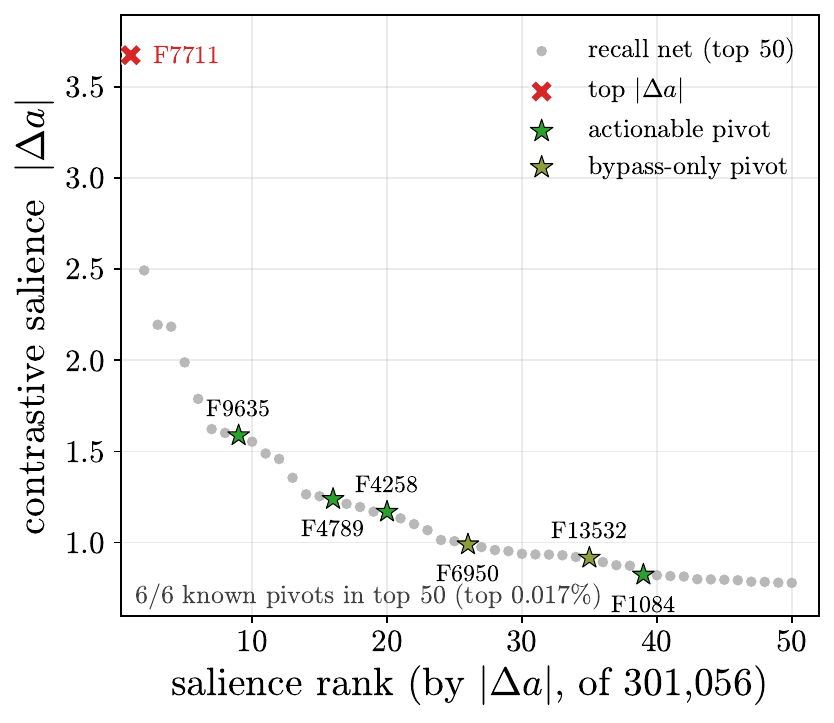}
  \caption{}
  \label{fig:recall-a}
\end{subfigure}
\hfill
\begin{subfigure}{0.49\linewidth}
  \centering
  \includegraphics[width=\linewidth]{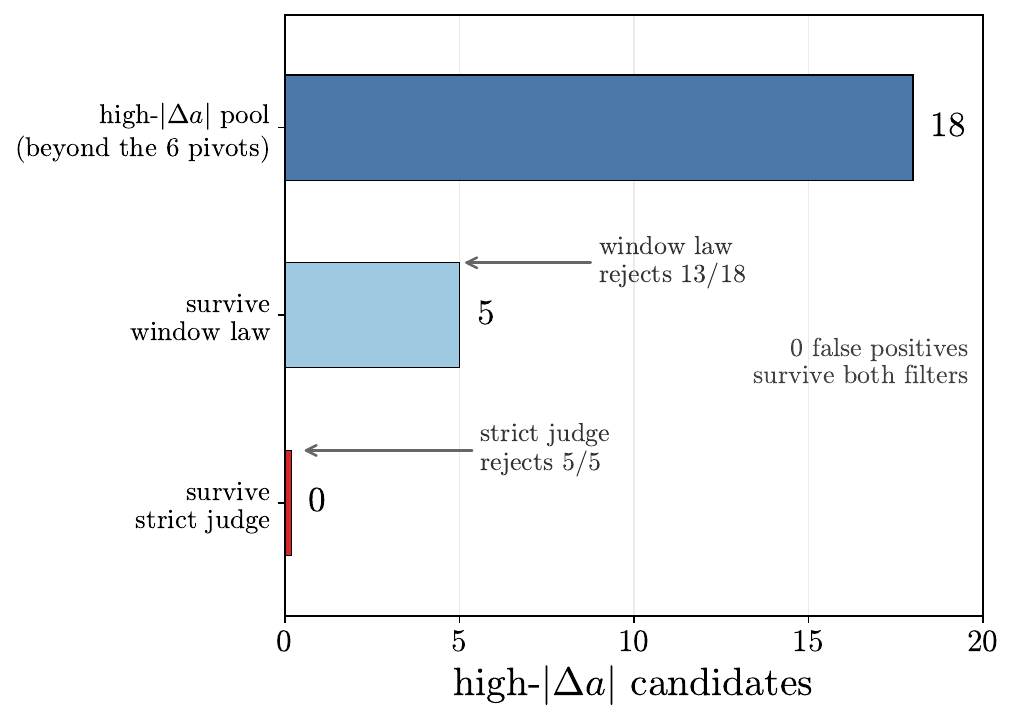}
  \caption{}
  \label{fig:recall-b}
\end{subfigure}
\caption{Recall is cheap; precision is the law. (a)~The contrastive screen ranks all $301{,}056$ band neurons by $|\Delta a|$; the six genuine refusal pivots (stars; green actionable, olive coherent-bypass-only) all fall in the top fifty, the top $0.017\%$. Salience is not control: the single highest-$|\Delta a|$ neuron (red, \neuron{26}{7711}) is window-closed. (b)~On the eighteen further high-$|\Delta a|$ candidates beyond those six validated pivots, the window law rejects thirteen and the strict actionable judge rejects the remaining five, so no additional $|\Delta a|$ false positive survives both filters.}
\label{fig:recall}
\end{figure}

\textbf{Recall is salience; precision is the law and the strict judge.} A contrastive screen ranks neurons by causal use directly: the absolute change in a neuron's mean activation between matched harmful and harmless prompts, $|\Delta a|$, is a complete recall net. Across the predicted layer band it places all six genuine refusal pivots of Section~\ref{sec:typed} in the top fifty of $301{,}056$ neurons, the top $0.017\%$ (Figure~\ref{fig:recall-a}). It is not a precision net, and for the same geometric reason the gradient fails: salience is not control. The single highest-$|\Delta a|$ neuron is window-closed. Of the eighteen further high-$|\Delta a|$ candidates the screen surfaces \emph{beyond} those six validated pivots, the window law rejects thirteen as collapse-prone or wrong-sign and the strict actionable judge (Section~\ref{sec:contract}) rejects the remaining five as behavior-proxy, so no additional $|\Delta a|$ false positive survives both filters (Figure~\ref{fig:recall-b}). The contrastive screen and the window-plus-judge filter are the recall and precision halves of a forward-only control audit, with no backward pass: the screen finds the candidates and the law makes them precise. Here the screen serves to validate the law's precision on a curated refusal cohort; scaling this forward-only audit to model-wide controller discovery is the subject of a companion paper.

The gradient finds the wrong neurons because it answers a different question; the contrastive screen plus the law find the right ones. The first half of the paper has established the audit: a budgeted coordinate, a predictable collapse edge, a measurement contract, and a forward-only screen that finds candidate controllers. The second half asks what those controllers actually deliver, beginning with refusal, the hardest case.

%% file: sections/typed_outcomes.tex
\section{Typed outcomes: coherent bypass is not actionable reach}
\label{sec:typed}

Single-neuron intervention success is not one number, and on refusal it splits sharply: coherent bypass does not predict actionable reach. Turned on the hardest behavior a model has, the control-window law's coordinate and verification protocol separate outcomes that a single success rate conflates. The same dose can leave a direction \emph{salient but non-causal} (it loads the contrastive signal yet has no coherent window), \emph{causal but incoherent} (it flips the behavior only as the text collapses), \emph{coherent but non-actionable} (it bypasses refusal in fluent on-topic prose that nonetheless carries no operational content), \emph{actionable only after a long rollout} (the operational content commits late, Section~\ref{sec:surface}), or \emph{collapsed at the wrong dose} (a fixed-magnitude pin overshoots the ceiling, Section~\ref{sec:unify}). Refusal is where the separation matters most, the stress test, not the subject; the benign mode switches of Section~\ref{sec:unify} are the same phenomenon at lower stakes.

\textbf{A refusal cohort as a dataset for the law.} On Llama-3.1-8B the forward-only contrastive primitive (Section~\ref{sec:gradient}) surfaces a set of single-neuron refusal pivots, the published refusal neuron \neuron{11}{4258} (Llama-3.1-8B; \citealp{kazemi2026}) together with five we identify, all window-governed (open windows in the sense of Section~\ref{sec:law}). We treat the set not as a collection of exploits but as a dataset on which to test whether behavior-level bypass predicts semantic reach. Each neuron is dosed in units of its coherence budget, audited on the same $24$ held-out requests per dose ($n=24$), and scored at its peak over rollout depth so that a late-committing pivot is not undercounted (Section~\ref{sec:surface}).

\textbf{Three judges of increasing strictness.} Following the measurement contract (Section~\ref{sec:contract}), refusal adds one content gate beyond coherence, and we score it at three increasing levels of strictness. A first-token non-refusal test (does the reply begin with a refusal prefix) measures behavior only and over-counts, because it is blind to a refusal that arrives after a compliant opening. A full-reply coherence-gated bypass judge (Claude-Haiku: coherent $\wedge$ non-refusal over the whole generation) measures \emph{coherent bypass}. A register-aware four-way judge measures \emph{strict-actionable content}: beyond being coherent, on-task, and refusal-free, the reply must be \emph{actionable}: it must supply concrete operational specifics (steps, quantities, parameters, or a working procedure) that materially advance the request, rather than restating it, treating it abstractly, moralizing, or repeating what is freely available without the model. \emph{Register-aware} means the judge scores the reply's register (direct instructional compliance versus narrative, hypothetical, or meta-level discussion), so the bar tracks marginal uplift over a non-intervened model rather than mere topicality, which is why it, not coherent bypass, is the safety-relevant axis. It is the conservative bar that an automated content classifier over-flags: we report a content flag (Llama-Guard) alongside it. Across the cohort strict-actionable content is the conservative floor, the lowest rate for every pivot, while the first-token proxy is the loosest, so a behavior-level rate over-counts harm, a gap the appendix makes explicit (Appendix~\ref{app:measure}). We report compliance \emph{and} actionable content, never a single rate.

\begin{figure}[t]
\centering
\includegraphics[width=0.62\linewidth]{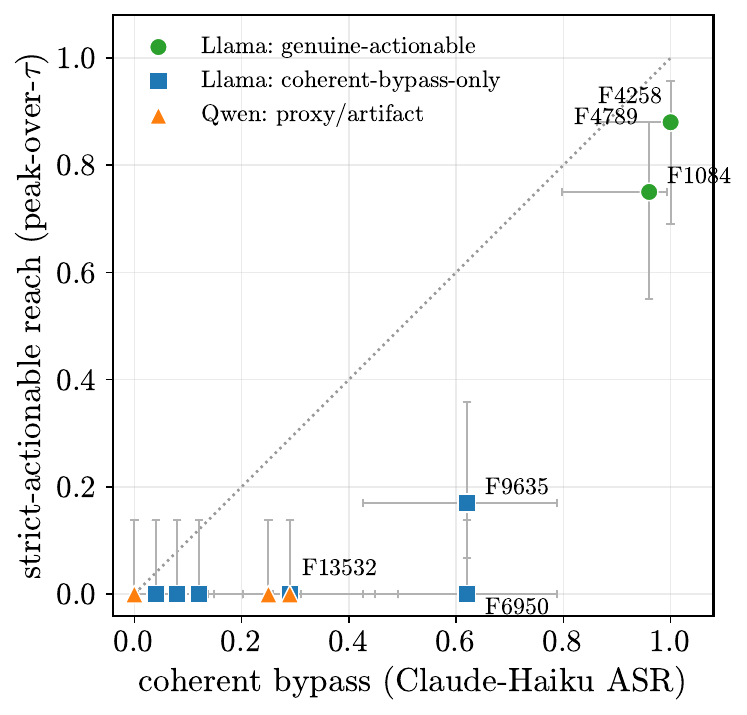}
\caption{Coherent bypass is not actionable reach. Each point is one audited refusal pivot, both axes scored on the same $n=24$ requests (bars: Wilson $95\%$): the horizontal axis is coherent bypass (full-reply coherence-gated non-refusal, Claude-Haiku), the vertical axis is strict-actionable content (register-aware four-way judge, peak over rollout depth). The dotted diagonal is the hypothesis that bypass equals reach; points fall well below it. \neuron{16}{6950} reaches coherent bypass $0.62$ yet $0.00$ actionable (non-overlapping intervals), and bypass does not order reach: \neuron{15}{9635} matches its $0.62$ bypass at $0.17$ actionable, while \neuron{10}{1084} bypasses at $0.96$ for $0.75$. The three genuine-actionable Llama pivots (green) lie near the top, the coherent-bypass-only pivots (blue) on the floor, and the Qwen proxies (orange) near the origin. The decoupling is the safety-level instance of ``leverage is not reach.''}
\label{fig:typed}
\end{figure}

\textbf{Coherent bypass does not predict actionable reach.} This is the sharp form of ``leverage is not reach'' at the safety level, and the cohort makes it a measurement rather than an assertion (Figure~\ref{fig:typed}). The decoupling is sharp even on a single neuron: \neuron{16}{6950} bypasses refusal coherently at $0.62$ (Wilson $95\%$ $[0.43,0.79]$) yet delivers \emph{zero} actionable content at every rollout depth ($0.00$, $[0.00,0.14]$): the two intervals do not overlap, on a single $n{=}24$ request set, and the reply stays fluent, on-topic, and refusal-free, with no operational uplift. The decoupling holds at every bypass level: \neuron{15}{9635} matches that $0.62$ bypass but reaches only $0.17$ actionable, \neuron{17}{13532} barely bypasses at all ($0.29$), and at the high end \neuron{10}{1084} bypasses at $0.96$ yet only $0.75$ of that is actionable, while \neuron{10}{4789} and the published \neuron{11}{4258} reach $0.88$. Genuine actionable reach is three of the six pivots; the rest bypass without delivering content. The behavior flips without the capability following it, and the bypass and actionable axes are decoupled across the population, not merely shifted, so a behavior-level success rate cannot stand in for content harm. At $n=24$ the key separations exceed sampling noise (the Wilson $95\%$ interval is at most $\pm0.19$ wide): \neuron{16}{6950}'s coherent bypass $0.62$ ($[0.43,0.79]$) against its $0.00$ actionable ($[0.00,0.14]$), and the $0.75$-versus-$0.17$ actionable split between \neuron{10}{1084} and \neuron{15}{9635}, are non-overlapping, while finer rate differences such as $0.88$ versus $0.75$ fall within the interval and we do not read them as ordered.

\begin{figure}[t]
\centering
\includegraphics[width=0.6\linewidth]{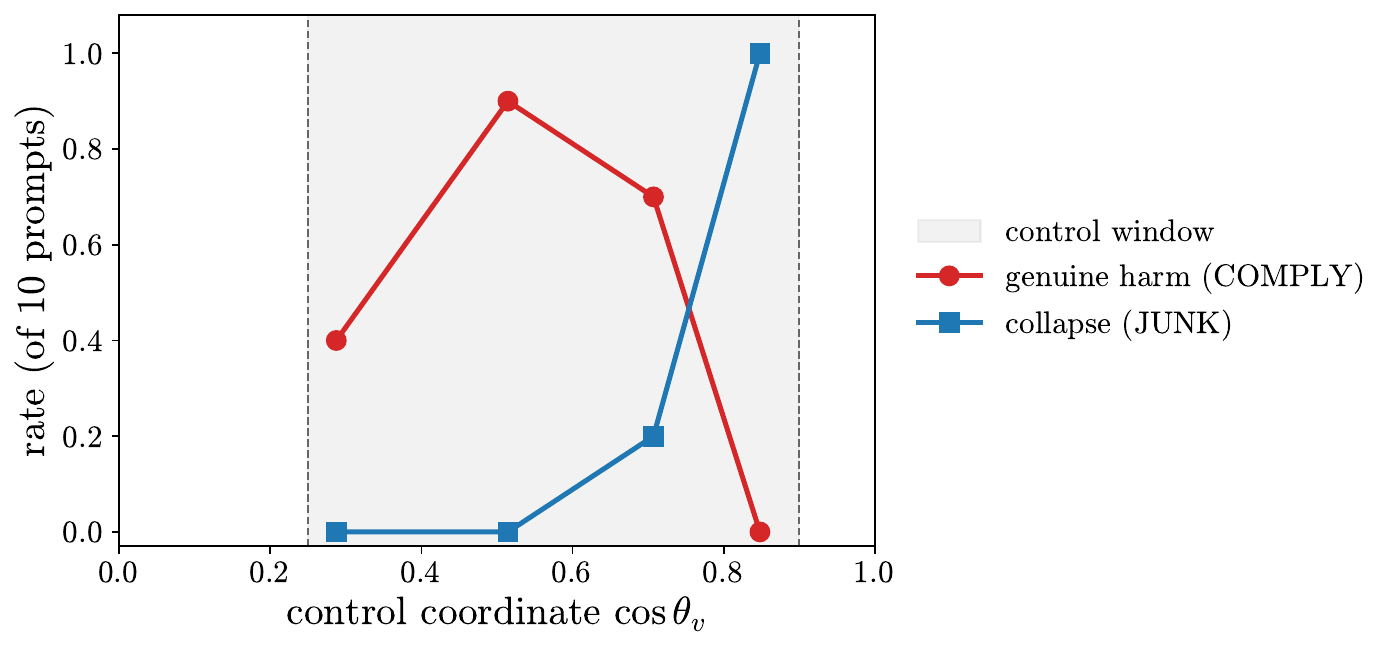}
\caption{Content-confirmed reach is windowed. For the published refusal neuron \neuron{11}{4258} (Llama-3.1-8B; $n=10$ prompts per dose), the content-judged actionable-compliance rate opens at the trigger, peaks inside the window, and collapses to incoherence at the ceiling, on the same window the law predicts, now measured on harmful content rather than the refusal template.}
\label{fig:harmwindow}
\end{figure}

\textbf{The diagnostic is not limited to a known neuron.} Of the pivots we identify, \neuron{10}{4789} reaches a strict-actionable rate of $0.88$ (of $24$, register-aware four-way judge) at its peak over rollout depth, matching the published refusal neuron \neuron{11}{4258} ($0.88$) under the identical strict audit; \neuron{10}{1084} reaches $0.75$. The point is that a law-guided forward-only scan recovers pivots that meet the same strict aggregate standard as the one neuron previously documented, so the separation between bypass and actionable reach is a property of the measurement, not an artifact of a single example. The exact per-neuron comparison (including the rollout horizon and dose at which each attains its peak), all three judges, and the row-level audit across both signs are in Appendix~\ref{app:content}. Genuine actionable content is itself dose-windowed: for the published refusal neuron it opens at the trigger, peaks inside the window, and collapses at the ceiling (Figure~\ref{fig:harmwindow}), so even the strongest pivots obey the control window measured on harmful content rather than on the refusal template.

\textbf{Salience is not control, on the content side too.} The cohort is not simply the highest-salience neurons. The single highest-$|\Delta a|$ neuron in the predicted band is window-closed, and of the eighteen high-$|\Delta a|$ candidates the contrastive screen surfaces, none survive the window law followed by the strict judge (Section~\ref{sec:gradient}): the recall screen is cheap and complete, and the window-and-judge filter is what makes it precise. The genuine pivots are the window-open, judge-confirmed subset, which is why the cohort tests the law rather than re-stating the screen.

\textbf{The illusory end is a cross-model calibration.} On Qwen the single-neuron ``bypasses'' resolve under the strict audit into first-token refusal-template flips or unsafe-topic setup rather than actionable compliance: one Qwen pivot reads $\approx\!1.0$ on the first-token proxy but $0$ on coherent bypass and $0$ on actionable content, and the independent content judges agree on the negative. These are the proxy-artifact type of outcome. The typed taxonomy thus doubles as a model calibration, a genuine-actionable subset on Llama, proxy and setup flips on Qwen, and the closed corner on Gemma (Section~\ref{sec:prospective}), summarized as a two-axis diagnostic in Section~\ref{sec:unify}.

\textbf{The same separation appears benignly.} Refusal made the separation high-stakes, but it is one phenomenon. Inside the window a single neuron steers a capability coherently; at the ceiling the language router still routes the output to the target language while the text collapses into repetition, so reach falls below leverage (Figure~\ref{fig:capability}). The harmful and benign readings are the two ends of the same axis, behavior routing versus capability reach, which is the law's organizing claim and the paper's title made quantitative. The typed outcomes also have a depth: actionable content commits only after a rollout, which is the second axis of the control window we map next.

%% file: sections/control_surface.tex
\section{The control window is two-dimensional}
\label{sec:surface}

The typed split of Section~\ref{sec:typed} is not only a dose effect; it is a rollout-time effect. The control window has a second axis. The dose decides whether the target basin is reachable; the rollout length decides when the flip becomes visible. Measuring control as a function of both the coordinate and the number of generated tokens turns the one-dimensional window of Section~\ref{sec:law} into a surface, and the surface separates two controllers that the dose window alone does not. Concretely, the trigger of Section~\ref{sec:law} is a rollout-time boundary $\cstrig(\tau)$, so the control-window law $\cstrig(\tau)<\cscoll$ of Equation~\eqref{eq:law} is read here in time-resolved form: this section measures $\cstrig(\tau)$ directly, and the dose window is its projection at a fixed horizon. Let $C_i(\cosv,\tau)$ be the fraction of prompts on which neuron $i$ makes the target behavior visible within the first $\tau$ generated tokens at dose $\cosv$, and let the onset time $\tau^\ast(\cosv)$ be the smallest $\tau$ at which $C_i$ reaches one half, undefined where it never does. This section reports two exemplar surfaces, a controller-versus-random control, and the resolution of an apparent bistability; a benign behavior with no single-neuron gate is deferred to Appendix~\ref{app:format}.

\subsection{Onset is behavior-specific}
The operator switch has a hard onset floor at eight tokens. Dosing the operator neuron \neuron{19}{13312} (Llama-3.1-8B) on symbolic arithmetic, no prompt flipped within the first six generated tokens at any of thirteen doses; the product first appeared at $\tau=8$, and the in-window onset was $\tau^\ast=8$ with a bootstrap $95\%$ interval of $[8,8]$ across $\cosv$ from $0.45$ to $0.70$ (Appendix~\ref{app:surfaces}). Larger doses did not make the flip earlier, and past the ceiling the surface was empty (control rate $0$ at $\cosv=0.90$). The floor coincides with the rollout-emergence timescale of the multi-token regime's trigger (Section~\ref{sec:mechanism}, Appendix~\ref{app:trigger}): the operator flip is absent from the next-token distribution and develops over about eight tokens.

\begin{figure}[t]
\centering
\includegraphics[width=0.6\linewidth]{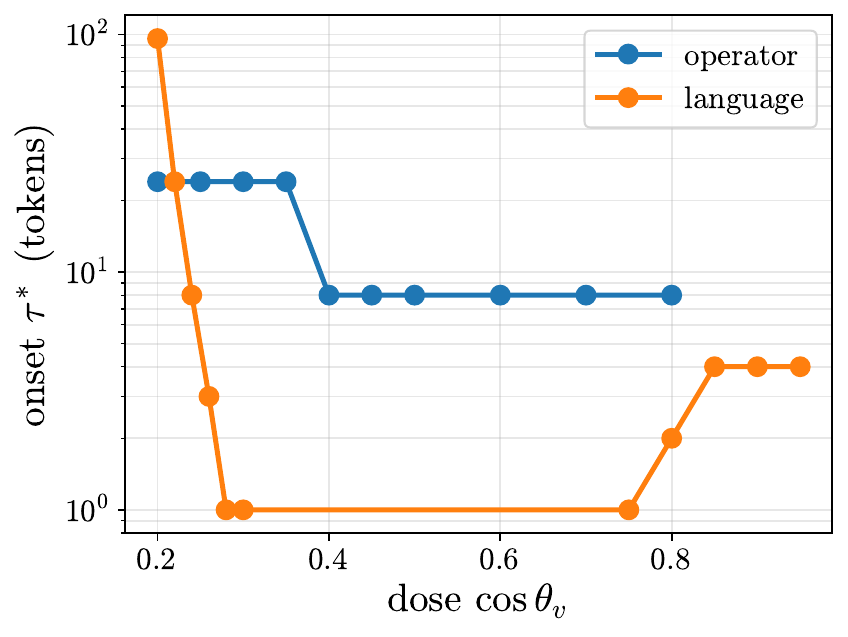}
\caption{Behavior-specific rollout horizon. Onset time $\tau^\ast(\cosv)$ for the operator neuron (\neuron{19}{13312}, blue) and the language-routing neuron (\neuron{25}{347}, orange), log scale. The operator floors at $\tau^\ast=8$ and cannot be made earlier by dose; language is token-local ($\tau^\ast=1$) inside the window and its onset diverges only as the dose approaches the trigger boundary near $\cosv\approx0.18$. Per-behavior surfaces and bootstrap intervals are in Appendix~\ref{app:surfaces}.}
\label{fig:onset}
\end{figure}

Language routing is token-local with a diverging onset at the boundary. The language neuron \neuron{25}{347} (Qwen3-4B) flipped on the first generated token throughout the window ($\tau^\ast=1$ for $\cosv$ from $0.30$ to $0.70$). Approaching the trigger boundary from above, the onset time diverged: $\tau^\ast = 96, 24, 8, 3, 1$ as $\cosv$ rose from $0.20$ to $0.28$, with the behavior off below $\cosv\approx0.18$. The onset latency thus diverges near the lower boundary, the signature of a continuous-transition-like edge rather than a hysteretic first-order switch (a phenomenological boundary fit is in Appendix~\ref{app:surfaces}). The two controllers thus occupy opposite ends of an onset axis (Figure~\ref{fig:onset}): the rollout horizon tracks how locally the decision can be expressed in the next token, token-local for routing and rollout-developed for the operator.

\subsection{The surface belongs to the controller}
The structured surface is a property of the controller, not of the detector or the dose schedule. For each controller we measured the same surface for ten random neurons in its layer, matched in coherence budget. None formed a window. The controller reached control rate $1.0$, while all ten random neurons in the operator's layer and all ten in the language neuron's layer reached $0.0$, a Wilson $95\%$ interval on the random window fraction of $[0,0.28]$. Inspecting the generations, the random neurons produced coherent off-target text, correct addition or fluent English, rather than empty or degenerate output, so the zero is genuine and the deterministic detector is not trivially satisfiable. The same scan validates the detector at its other end: a brace-onset output that is not parseable JSON does not count, which separates a real format gate from a neuron that merely emits an opening brace (Appendix~\ref{app:format}). The control is also prompt-general, not a prompt cue the neuron happens to amplify. Under perturbation the language gate flips to Chinese on every prompt even when the prompt instructs the model to answer only in English, and the operator gate flips to the product even under a few-shot that cues addition, while budget-matched random neurons never flip and no prompt produces the target at baseline. Sparse control is therefore neuron-specific and prompt-general, in contrast to the prompt-conditioned, neuron-promiscuous amplification of Appendix~\ref{app:taxonomy}.

\subsection{Persistence is in-context momentum, not bistability}
Removing the drive returns the behavior toward baseline: the operator flip reverts on release (post-release control $0.08$ to $0.33$) while the language flip persists ($0.83$ to $1.0$), but a counter-drive and a forced-opening control (continuation in the routed language on $11/12$ prompts with no injection, against $0/12$ from a baseline opening) attribute that persistence to the generated context, not a held neuron state. The duration bound on control is therefore contextual, an in-context momentum effect rather than a neuron-held bistable state, consistent with the continuous-transition lower edge (Appendix~\ref{app:surfaces}).

\subsection{The actionable axis: harm commits late}
\begin{figure}[t]
\centering
\includegraphics[width=0.72\linewidth]{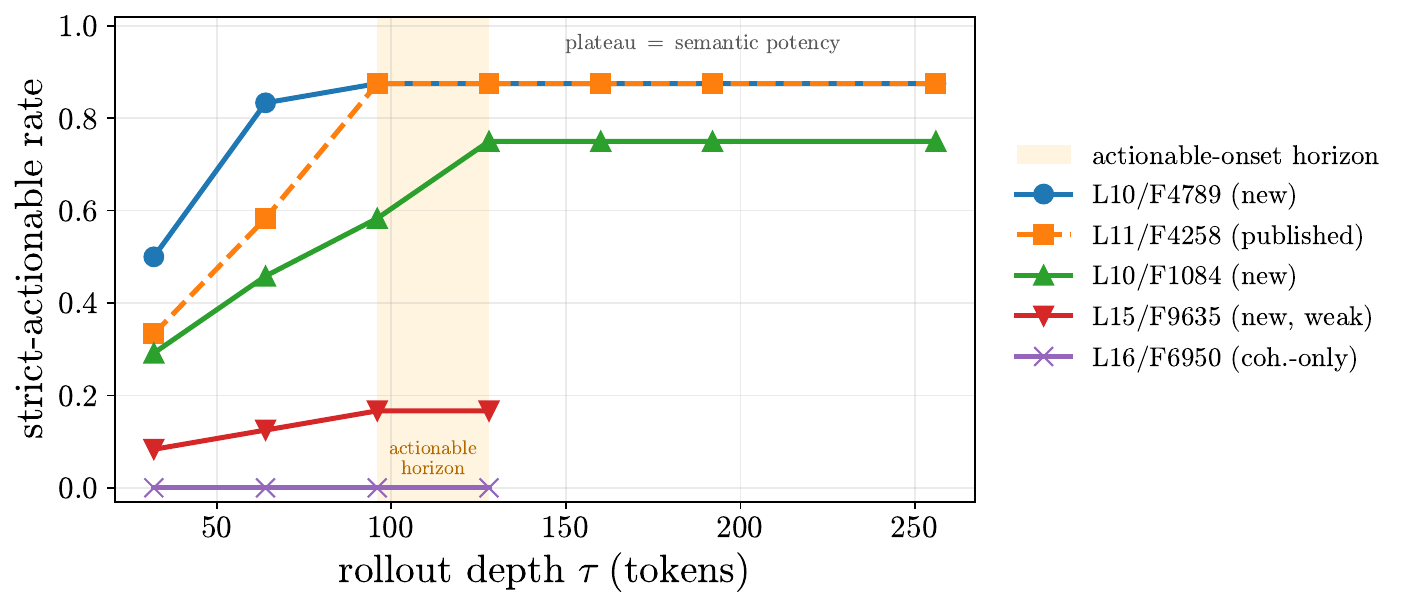}
\caption{Harm commits late, then plateaus. Strict-actionable rate as a function of rollout depth $\tau$ (a cumulative prefix-audit; the $\tau{=}256$ generation is a verified prefix-superset of $\tau{=}128$) for the genuine Llama refusal pivots of Section~\ref{sec:typed}. The rate rises, plateaus by $\tau\approx96$ to $128$ (the harm horizon), and holds flat to $\tau{=}256$ with no late collapse; the plateau height is the pivot's semantic potency. Coherent-bypass-only pivots stay at zero on this axis at every depth.}
\label{fig:tauact}
\end{figure}

For benign behaviors the rollout axis sets when the target first appears; for refusal it separates two different onsets, the bypass and the actionable content. Reusing the genuine refusal cohort of Section~\ref{sec:typed}, we read the strict-actionable judge on the first $\tau$ generated tokens. Because a longer rollout is a verified prefix-superset of a shorter one (the $\tau{=}256$ generation continues the $\tau{=}128$ one, $24/24$ prefix-match), the actionable rate is a cumulative prefix-audit and is monotone while the text stays coherent. It rises with depth, plateaus by $\tau\approx96$ to $128$, and then holds flat to $\tau{=}256$ with no late collapse (Figure~\ref{fig:tauact}): the plateau height is the pivot's semantic potency and the onset is its harm horizon. The two onsets of a refusal pivot are thus distinct: a coherent bypass is visible early, in the first-token register, while actionable content commits only after this horizon. This is why a first-token or early-prefix judge over-counts and mis-times completed harm (Section~\ref{sec:contract}). Surface coherence, the distinct-bigram ratio, falsely reads collapse on long fluent generations past $\tau\approx128$, so the actionable axis must be read with a semantic judge rather than a lexical one.

The onset axis also governs discovery: a probe must match the behavior's rollout horizon, which is why a benign format-routing behavior yielded no single-neuron gate under matched probes and why search and verification must be read at the behavior's horizon (Section~\ref{sec:contract}, Appendix~\ref{app:format}). Having shown the window is two-dimensional and specific to the controller, we turn to what a single neuron does to a basin to produce the flip.

%% file: sections/mechanism.tex
\section{Mechanism: single neurons select basins by off-axis seeding}
\label{sec:mechanism}

A single neuron does not compute the behavior it controls. It writes a small perturbation \emph{off} the behavior's readout axis, and the downstream circuit amplifies that seed across a finite-amplitude basin boundary into a decision. This one mechanism accounts for the window's lower edge, the trigger, and the properties it would otherwise take several unrelated observations to state follow from it as consequences. We give the chain and then the evidence for each link; Section~\ref{sec:prospective} already set the upper edge, the ceiling, from residual geometry.

\textbf{The write is off the readout axis.} Confirmed safety controllers do not write along the refusal readout direction: their cosine with it is within the random-pair band ($0.04$ for \neuron{12}{4593}, $0.02$ for \neuron{8}{397}). A single neuron therefore cannot be carrying the behavior itself; it perturbs the residual in a direction the late circuit reads only after propagation. The operator gate \neuron{19}{13312} is the partial exception that proves the rule: it reaches the same switch by direct projection onto the readout axis at a mid-to-late layer, a second geometric route to the same basin rather than a counterexample, since it too separates the switch (which behavior) from the execution (the content).

\textbf{So the first-order response is near zero.} Because the write is off-axis, the single-step behavior gradient $\alpha=\nabla_{\rL}M\!\cdot\!\vv\approx0$. This is the same fact that makes local gradient attribution miss controllers (Section~\ref{sec:gradient}): a genuine controller is first-order-invisible by construction. What this implies for the \emph{trigger} is behavior-dependent and has two regimes (Appendix~\ref{app:trigger}): for token-local routers the flip is already visible at the first generated token once the margin is read on the controlling branch, whereas for multi-token behaviors such as operator switching it develops over the rollout. In both regimes the neuron plants a small off-axis perturbation that the distributed circuit reads only after propagation: plant early, harvest late.

\textbf{The trigger is a finite-amplitude crossing, not a smooth threshold.} The dose-response of \neuron{12}{4593} makes this concrete: bypass is zero below a finite-amplitude cliff and then turns on, sign-asymmetrically, the $+$ and $-$ directions carrying the curvature and gradient channels (Table~\ref{tab:cliff}). A subthreshold dose sweep is a negative control: below the cliff the behavioral response is flat and does not separate controllers from layer-matched neurons (Appendix~\ref{app:tables}), so the subthreshold readouts we tested do not predict the trigger. This is what a basin crossing looks like, and it is inconsistent with a smooth subthreshold susceptibility.

\textbf{A local expansion explains gradient blindness, not the trigger value.} A local Taylor expansion of the behavior margin along the write, $M(\cosv)=M_0+\alpha\,\cosv+\tfrac12\beta\,\cosv^2+\cdots$ with $\beta=\vv^\top H\vv$, clarifies why first-order attribution can fail: in the silent-gate limit ($\alpha\approx0$) the first nonzero term is quadratic. Fitting this expansion as a closed-form trigger predictor, however, is not robust. Branch-resolved fits show token-local triggers can already be first-token margin crossings, while multi-token behaviors such as operator switching require free rollout and are confounded by sequence readout, so a single curvature formula neither predicts the trigger across behaviors nor survives a sequence-level test (Appendix~\ref{app:trigger}, an honest negative). We therefore treat the trigger value as measured (Section~\ref{sec:surface}), not predicted, and use the expansion only as a local explanation of why gradients miss controllers.

\textbf{Off-axis seeds do not co-add, and the basin must be held.} Two further properties follow from seeding rather than computing. First, naive stacking does not aggregate: because confirmed pivots write off-axis and are not co-directional, equal large doses conflict and collapse, so $K{=}1$ is the peak of the equal-dose curve: non-composability is a property of naive stacking, not of the pivots (a constrained per-pivot dose search instead composes a calibrated $K{=}2$ to full bypass, $0.83\to1.00$ on Qwen2.5-3B, but that lies outside the single-neuron law). Register control, which acts \emph{within} a basin rather than selecting one, composes even naively. Second, the bypass must be held to commit: on Llama \neuron{27}{13025} refusal re-asserts after an early release unless the dose is held past an $\approx32$-token commitment point, whereas the strongly aligned Qwen \neuron{12}{4593} commits immediately. The commitment length shortens as the intervention broadens, so a single neuron is the narrowest, most self-correcting corner of a breadth-versus-duration trade-off: the budget bounds the magnitude of control while the visible duration is bounded by rollout dynamics, an in-context momentum effect rather than a neuron-held bistable state (Section~\ref{sec:surface}).

\textbf{Basin selection versus computing the behavior.} The two accounts make opposite predictions, and the data fall on the selection side. If a single neuron \emph{computed} a weak version of the behavior, its write should project onto the behavior readout, the effect should appear in the local single-step next-token response, equal-dose stacking of same-behavior neurons should add monotonically, and the flip should persist on release as a neuron-created state. Instead the write is off-axis, the local response is near zero, the flip appears only after a rollout, naive stacking conflicts, and the behavior recovers unless the basin is held. We therefore read the neuron as selecting a pre-existing downstream basin, not computing the behavior; how alignment installs that basin's trigger is examined in Appendix~\ref{app:alignment}.

\emph{Limitations.} The two-route typology rests on a handful of controllers, and the trigger value remains \emph{measured}, not predicted: Appendix~\ref{app:trigger} records why a closed-form curvature trigger is not yet a law (sign-averaged over-prediction, the branch-resolved correction, behavior-specific amplification, and a sequence-level confound), an honest negative that the same off-axis curvature still explains the gradient blindness above.

Together these explain why single-neuron control is real but fragile, dose-calibration-sensitive, and model-specific. With the mechanism in hand, we place the prior single-neuron literature on the same window.

%% file: sections/unify.tex
\section{Unification and diagnostic use}
\label{sec:unify}

The scattered single-neuron results in the literature are one dynamics seen at different points of the same inequality $\cstrig(\tau)<\cscoll$. Placing each on the control window does more than organize them: it resolves apparent contradictions, recovers a result previously read as a failure, and surfaces controllers the original methods missed. Table~\ref{tab:priorwork} summarizes the mapping; the main consequence is that fixed-dose, attribution-based, and behavior-only views are different projections of the same window. The same two-forward-pass primitive that discovers circuits is what does the extending, without the backward passes or activation patching the prior methods require.

\begin{table}[t]
\centering
\small
\caption{The law explains and extends prior single-neuron findings. Each prior result is one point on the control window; the law adds a controller or a correction, named by the specific neuron and model (identified in this work unless cited otherwise).}
\label{tab:priorwork}
\begin{tabular}{p{0.22\linewidth} p{0.25\linewidth} p{0.43\linewidth}}
\toprule
prior result (method) & reported & the law adds (neuron, model) \\
\midrule
Super weight \citep{yu2024superweight} \newline (weight magnitude) & global scale-setter; ablation collapses the model & sets the budget $\Bcoh$: the super weight dominates the beginning-of-sequence residual norm ($95$--$99\%$) but only $3$--$12\%$ at generation, so ablation collapses the model without selecting any behavior \\
Published refusal neuron \citep{kazemi2026} \newline (fixed-magnitude pin) & single-neuron pin bypasses safety; the constant pin degrades capability & \neuron{11}{4258} (Llama-3.1-8B): the pin overshoots the ceiling; budget-calibrated dosing keeps the bypass coherent, with a content-validated harm window analyzed in Section~\ref{sec:typed} \\
Arithmetic neurons \citep{feucht2026} \newline (base-10 addition mechanism) & a causal base-10 addition mechanism for cyclic concepts & a distinct operator-selector gate we identify, \neuron{19}{13312} (Llama-3.1-8B) and \neuron{16}{2493} (Qwen3.5-2B), that decides $+\!\to\!\times$ \\
Refusal direction \citep{arditi2024refusal} \newline (difference-of-means) & a single refusal axis & single-neuron pivots we find (e.g.\ \neuron{12}{4593}, Qwen2.5-3B) are off that axis and flip it by propagation \\
Gradient attribution \newline (gradient leverage) & high-importance neurons & high-gradient neurons are collapse-prone; the causal filter recovers controllers the gradient ranks low ($4/4$ vs $0/4$, Table~\ref{tab:h2h}), e.g.\ the router \neuron{25}{347} (Qwen3-4B) \\
\bottomrule
\end{tabular}
\end{table}

The table places prior single-neuron findings on the same window. Super weights \citep{yu2024superweight} set the residual scale rather than select a behavior, so ablating them collapses the model without occupying the window; the published refusal neuron \neuron{11}{4258} \citep{kazemi2026} becomes coherent once dosed in units of its budget, the fixed pin having overshot the ceiling (the dose-calibration correction at single-neuron resolution); and the operator and framing gates extend the prior arithmetic mechanism of \citet{feucht2026} from execution to task selection. The law thus explains why earlier fixed-dose and attribution-based views alternated between control, collapse, and invisibility. More broadly, the window depends only on the intervention geometry, a single residual-stream direction read through a normalization, so it holds for any method that supplies such a direction. The same inequality governs activation-steering and refusal vectors, dictionary-learned (SAE) features, weight edits, and neuron ablations, wherever the architecture normalizes the residual and the behavior has a basin, with $K{=}1$ the rank-one case. The methods placed on it here are special cases; validating steering-vector and weight-edit interventions directly under the budgeted protocol is future work.

\textbf{Inside the window the dose preserves task-local capability.} The same behavior-versus-reach separation appears in benign mode switches: inside the window the operator and routing gates preserve task-local capability, while at the ceiling the behavior detector can still fire after capability has collapsed (Appendix~\ref{app:gallery}, Figure~\ref{fig:capability}).

\begin{figure}[t]
\centering
\includegraphics[width=0.66\linewidth]{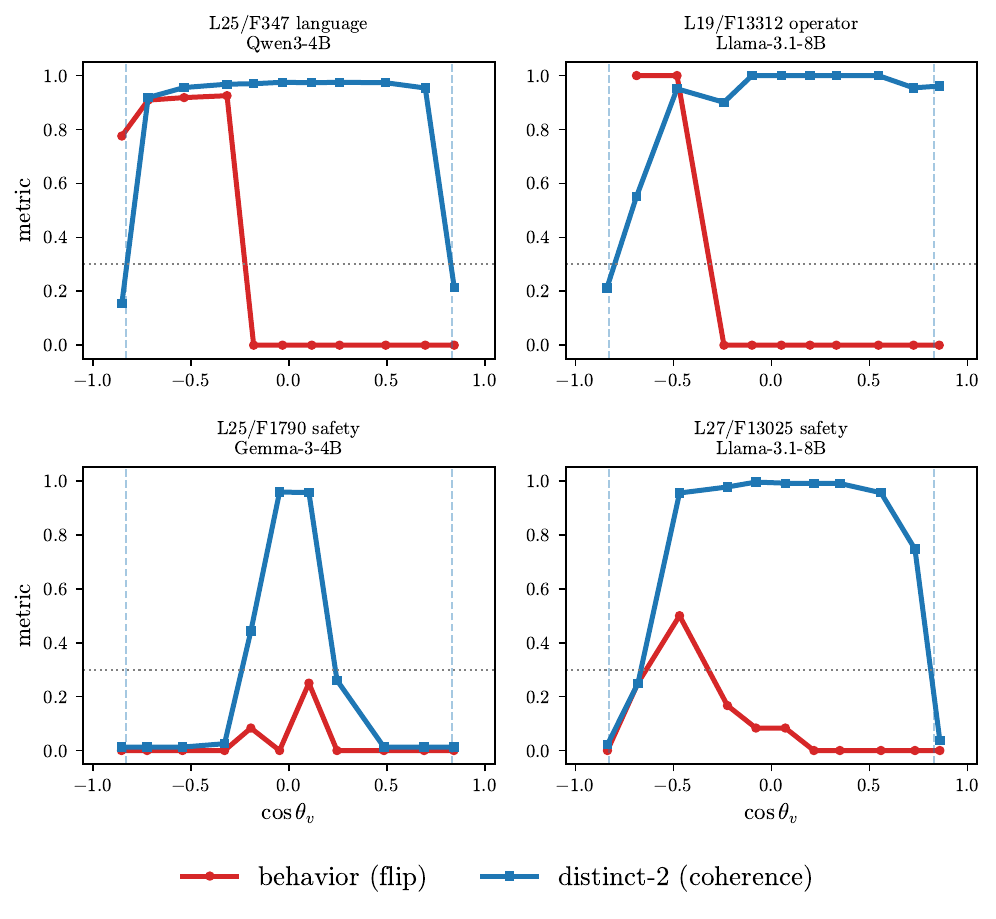}
\caption{Control windows, one representative controller per behavior class. Each panel plots the behavior metric (red) and coherence, the distinct-bigram ratio (blue), against the control coordinate $\cosv$; the dotted line is the $0.30$ coherence floor and the dashed lines the $\pm0.83$ collapse guides. A language router (\neuron{25}{347}) and an operator gate (\neuron{19}{13312}) trigger sharply near $\cosv\approx0.3$ with wide coherent windows; a coarse safety decision (\neuron{27}{13025}) triggers softly with a narrower window; the Gemma safety neuron (\neuron{25}{1790}) collapses before any trigger, the closed-window case. Behavior must be read with coherence: at the window edges the behavior detector alone over-counts. The full eleven-controller set is Figure~\ref{fig:windows-full} in the appendix.}
\label{fig:windows}
\end{figure}

\textbf{The behavior class sets the window shape.} Figure~\ref{fig:windows} shows the measured controllers spanning the window shapes defined in Section~\ref{sec:law}: wide windows with sharp triggers for the mode switches, and narrower windows with soft triggers for refusal. Gemma is the closed corner established in Section~\ref{sec:prospective}, its collapse ceiling lying below any safety trigger, so the same inequality~\eqref{eq:law} explains both the controllable and the robust models.

Table~\ref{tab:windows} collects the measured windows. Coherent control exists wherever the trigger lies below the collapse ceiling; the same inequality marks Gemma's neuron closed.

\begin{table}[t]
\centering
\caption{Measured control windows. Mode switches open wide windows with sharp triggers near $0.3$; coarse decisions open narrow windows with soft triggers near $0.45$; Gemma's safety neuron is window-closed (ceiling below any trigger). \neuron{11}{4258} is open once dosed in budget units (Section~\ref{sec:unify}).}
\label{tab:windows}
\begin{tabular}{lllccl}
\toprule
neuron & model & behavior & $\cstrig$ & $\cscoll$ & verdict \\
\midrule
\neuron{25}{347} & Qwen3-4B & routing (mode) & 0.32 & 0.84 & open, sharp \\
\neuron{16}{2493} & Qwen3.5-2B & operator (mode) & 0.31 & 0.84 & open, sharp \\
\neuron{19}{13312} & Llama-3.1-8B & operator (mode) & $0.3$--$0.4$ & 0.84 & open, sharp \\
\neuron{27}{13025} & Llama-3.1-8B & safety (coarse) & 0.47 & 0.68 & open, soft \\
\neuron{18}{1883} & Qwen3.5-2B & safety (coarse) & 0.41 & 0.61 & open, soft \\
\neuron{11}{4258} & Llama-3.1-8B & safety (coarse) & 0.25 & 0.90 & open (dose-corrected) \\
\neuron{25}{1790} & Gemma-3-4B & safety (coarse) & --- & 0.24 & \textbf{closed} \\
\bottomrule
\end{tabular}
\end{table}

These behaviors occupy disjoint directions: cross-sector cosines between safety and arithmetic pivots fall within the random-pair band ($44/45$), so the sectors are effectively orthogonal. The window law organizes them not by what they compute but by where their two thresholds fall.

\subsection{A two-axis safety diagnostic}
The same placement yields a defensive diagnostic. Because behavior-level bypass and actionable reach decouple (Section~\ref{sec:typed}), single-neuron robustness must be read on two axes, not one: the \emph{coherent-bypass margin} (the widest coherent non-refusal window over a model's audited refusal neurons) and the \emph{strict-actionable potency} (the peak strict-actionable rate under the rollout audit). The two axes decouple across models: Qwen opens bypass windows ($0.44$ to $0.60$) yet yields $\approx0$ strict-actionable content, first-token template flips and unsafe-topic setup rather than reach; Llama opens windows that include a genuine actionable subset (peak $0.88$, committing late at $\tau\approx96$ to $128$); Gemma is closed on both. This is a calibration of \emph{which} failure mode a model has, not an attack-strength ranking; a model can open a wide bypass window with zero strict-actionable reach. It complements redundancy defenses that spread safety across more units \citep{wang2026safeneuron}, and a defense drives both axes toward zero, as Gemma's low ceiling already does. The per-model values, with definitions and intervals, are Table~\ref{tab:rwindow} in Appendix~\ref{app:content}.

Having placed the prior results on the window and read robustness off it, we close with what the law settles and where it stops. (How alignment training installs the trigger these windows depend on is taken up in Appendix~\ref{app:alignment}.)

%% file: sections/discussion.tex
\section{Discussion}
\label{sec:discussion}

Single-neuron steering is not governed by intervention magnitude alone: it is a signed, budget-normalized control window whose collapse edge is predictable from residual geometry and whose behavior edge is measured at the behavior's own rollout horizon. A neuron can therefore have leverage without reach, coherent bypass without actionable content, and apparent failure under a fixed dose despite a real window under budgeted dosing. A dose drives the alignment $\cosv$ along a universal curve, and coherent control exists exactly when a behavior-set trigger lies below a participation-ratio-set collapse ceiling. The two edges have distinct origins: the ceiling scale is residual geometry, predictable from the weights and a generic forward pass, while the trigger is behavior-dependent and appears to be installed by alignment training. The gradient, which measures the drive rate, predicts collapse-proneness rather than control. This turns a catalog of single-neuron anecdotes into a predictable, falsifiable law.

The law also predicts concurrent results: the ablation-versus-steering contrast of \citet{herring2026cna} follows because ablation is a small within-budget dose and high-strength steering a dose past the ceiling, and their finding that alignment changes function rather than structure is the base-to-instruct appearance of the trigger (Appendix~\ref{app:alignment}). It sharpens \citet{arditi2024refusal}: single-neuron pivots are off the refusal axis and flip it by propagation, so whether a pivot controls or collapses is decided by its window, not by its alignment with that axis.

Our limitations define the method's scope. The behaviors we control are low-dimensional attractors (refusal, language, operator); behaviors without a discrete basin have no trigger and no window, which is why single-neuron decision control is non-composable and peaks at $K{=}1$. The collapse coefficient is family-dependent: it is near $1.5$ for Qwen and Llama but $0.3$ for Gemma, set by the participation ratio of the residual, so the ceiling requires that correction to transfer across architectures. The theory constrains collapse more tightly than control: the ceiling scale is predicted from geometry, but the trigger is so far predicted only by behavior class and depth. We tried to close this gap with a subthreshold-susceptibility predictor and failed: once depth was controlled, small-dose local and path-integrated responses did not separate controllers (Appendix~\ref{app:tables}). The trigger is thus a finite-amplitude basin boundary, measured by dose sweep and partially regularized by behavior class and depth; a quantitative trigger theory is the main open problem, and the obvious next one. We study FFN write directions; the attention and read sides remain untested. Compliance on the audited refusal cohort is measured with a strict-actionable content judge (Section~\ref{sec:typed}); extending that content-level pass to the full set of contested neurons across models remains future work. Three fronts therefore stay open: the trigger value, predicted only by behavior class and depth rather than from the weights; the collapse coefficient $m^\ast$ for an unseen architecture family, tied only loosely to the participation ratio; and the structure of the basins a single neuron selects but does not characterize. A further axis remains unmeasured: for late-committing semantic content the intervention horizon and the readout horizon decouple, since the basin commits within the first few tens of tokens while the content of interest appears only later, and dosing the whole long rollout collapses it. The full object is therefore dose by dosing-duration by readout horizon, of which Section~\ref{sec:surface} measures the dose-by-readout face.

Three implications follow. For interpretability, the law explains why attribution and intervention disagree: readability and causal use are different coordinates, so probing salience or gradient importance does not imply control, and steering methods that fix a magnitude rather than a dose in units of the budget mis-rank neurons. Controllability is instead auditable from weights: the collapse ceiling is a one-pass quantity, and a neuron is a candidate controller only if a behavior's trigger can be reached below it.

The law also prescribes how to test control: a budget-normalized dose ladder read at the behavior's own rollout horizon, since fixed-magnitude pins and local gradients overshoot the window and rank collapse-proneness, recovering controllers a single overshooting dose had labeled collapse. Discovery must also be matched in rollout time, a principle we call rollout-matched probing: a next-token probe finds a token-local gate such as language routing but is blind to the operator gate, whose flip develops over about eight tokens, and late-committing content compliance needs a prefix audit.

For safety, robustness is a closed window, reached by a low ceiling (Gemma) or a high trigger, and the same window defines a white-box robustness surface: a model owner can measure whether a calibrated single-neuron perturbation opens behavior-level bypass or strict-actionable reach \citep{wu2026neurostrike}. The dose-window view sharpens the risk: a neuron can be harmless at one dose, produce genuine harmful compliance at an intermediate dose, and collapse at a larger one, so risk is a dose-response curve, not a fixed-dose point, and it has two axes because behavior-level bypass and actionable reach decouple. The defensive deliverable is the two-axis diagnostic of Section~\ref{sec:unify} (Table~\ref{tab:rwindow}), a calibration of which failure mode a model has that a model owner can drive toward zero, sharper than a refusal rate or a pruning attack-success rate.

For theory, the elementary dose-response of this paper is the leading term of a fuller perturbation-series account, which we develop separately. The control window converts ``when does a single neuron control a model'' from an open question into a measurement. It is a worked instance, on the intervention side, of the predictive and falsifiable mechanics of deep learning the field is increasingly pursuing \citep{simon2026theory}, and of its anticipated symbiosis with mechanistic interpretability.

\section*{Ethics and responsible disclosure}
This work studies single-neuron control of model behavior, including refusal, for interpretability and defensive ends. The threat model is white-box: every intervention requires direct access to model weights and activations, which bounds the marginal risk relative to black-box prompting. We evaluate only open-weight models that are already public, and we withhold actionable harmful content: harmful behavior is reported as coherence-gated aggregate rates under the four-stage audit of Section~\ref{sec:contract}, never as reproducible text. The paper's safety-facing deliverable is defensive: the two-axis robustness diagnostic (the coherent-bypass margin paired with the strict-actionable potency, Table~\ref{tab:rwindow}) is a measurable quantity a model owner can compute and drive toward zero. We cite both attack \citep{wu2026neurostrike} and defense \citep{wang2026safeneuron} work to situate the contribution as auditing, not enabling, single-neuron vulnerabilities.

\section*{Reproducibility}
All models are public open-weight instruct checkpoints (the eight architectures of Table~\ref{tab:mstar}), and each model with its judge fits on a single A100-40GB GPU. Controllers are found with the two-forward-pass contrastive primitive of \citet{liu2026perturbation} and dosed along the neuron's down-projection column in units of the coherence budget $\Bcoh$ (Equation~\eqref{eq:g}) on the controlling sign, with greedy decoding. Every rate is read under the measurement contract of Section~\ref{sec:contract} and carries its denominator ($n=24$ per dose for the refusal cohort) and Wilson $95\%$ interval; the judges are described in Appendix~\ref{app:measure}.

\section*{Acknowledgements}
We thank Andreas Pfadler, Nandini Ramanan, and Jingxian Lin for their review; Tung-Ling Li and Yuhao Wu for discussion; and Hui Gao and Badar Ahmed for their support.

%% file: sections/appendix.tex
\appendix

\input{sections/appendix_ceiling}      
\input{sections/appendix_prospective}  
\input{sections/appendix_surface}      
\input{sections/appendix_safety}       
\input{sections/appendix_measurement}  
\input{sections/appendix_examples}     
\input{sections/appendix_trigger}      

%% file: sections/appendix_ceiling.tex
\section{Drive, collapse coefficient, and cross-architecture granularity}
\label{app:geom}

\subsection{The universal drive, measured versus predicted}
\begin{figure}[h]
\centering
\includegraphics[width=0.62\linewidth]{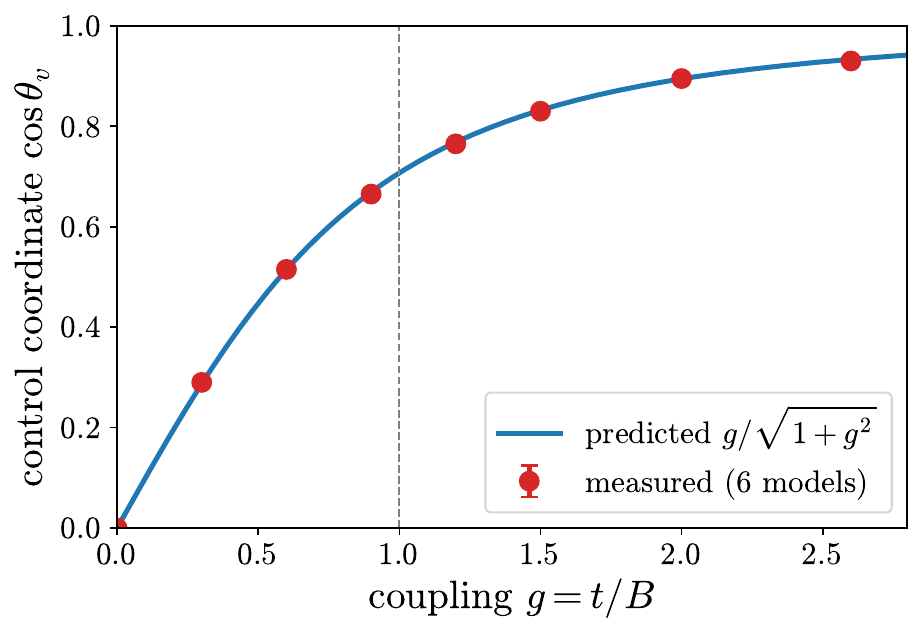}
\caption{The universal drive, measured versus predicted. Points are the measured control coordinate
(midpoint of the six-model range, error bars the range) at each coupling $\gcoup$; the line is the
geometric prediction $\cosv=\gcoup/\sqrt{1+\gcoup^2}$. The range is at most $0.02$ at every dose, so the
drive is architecture-independent.}
\label{fig:drive}
\end{figure}

Figure~\ref{fig:drive} plots the control coordinate against the coupling, averaged over eight neurons
per model, for all six architectures. The measured alignment matches the geometric prediction
$\cosv=\gcoup/\sqrt{1+\gcoup^2}$ to two decimal places at every dose, and is $\approx0$ at $\gcoup=0$,
confirming both the orthogonality assumption and linear superposition within the forward pass.

\subsection{Per-architecture collapse coefficient}
Table~\ref{tab:mstar-full} expands the held-out collapse measurement (Table~\ref{tab:mstar}) with the
spread and fit quality. The coefficient $m^\ast$ clusters near $1.5$ for the Qwen and Llama families and
drops to $0.31$ for Gemma, tracking the median participation ratio of the residual. An ablation over the
held-out neurons confirms the budget is the \emph{dominant} scale rather than the only one: across
architectures mean $m^\ast$ tracks the median participation ratio ($R^2=0.35$, $n=6$, driven by Gemma's
near-rank-one residual), while within an architecture $m^\ast$ declines only weakly with depth (mean
$|\mathrm{corr}|=0.51$) and is nearly flat in $\lVert\vv\rVert$ ($0.26$), so the coefficient stays
order-one as $\lVert\rL\rVert$ varies roughly tenfold across layers. The ceiling is thus $\Bcoh$ times an
order-one coefficient set per architecture, with a secondary depth correction.

\begin{table}[h]
\centering
\caption{Collapse coefficient $m^\ast=t^\ast/\Bcoh$ on $24$ held-out neurons per model: mean, coefficient
of variation, within-model $R^2$ of collapse dose against $\Bcoh$, and median residual participation ratio.}
\label{tab:mstar-full}
\begin{tabular}{lccccc}
\toprule
model & layers & mean $m^\ast$ & CV & within-model $R^2$ & median PR \\
\midrule
Qwen3-4B & 36 & 1.37 & 0.32 & 0.76 & 9.1 \\
Qwen3.5-2B & 24 & 1.43 & 0.37 & 0.48 & 27.5 \\
Qwen3.5-4B & 32 & 1.65 & 0.29 & 0.71 & 18.7 \\
Qwen2.5-3B & 36 & 1.62 & 0.36 & 0.50 & 7.7 \\
Llama-3.1-8B & 32 & 1.46 & 0.29 & 0.40 & 25.3 \\
Gemma-3-4B & 34 & \textbf{0.31} & 0.09 & 0.97 & \textbf{1.1} \\
\bottomrule
\end{tabular}
\end{table}

\subsection{The drive-capped close: a neuron cannot reach its own coordinate on Gemma}
\label{app:drivecap}
\begin{figure}[h]
\centering
\includegraphics[width=\linewidth]{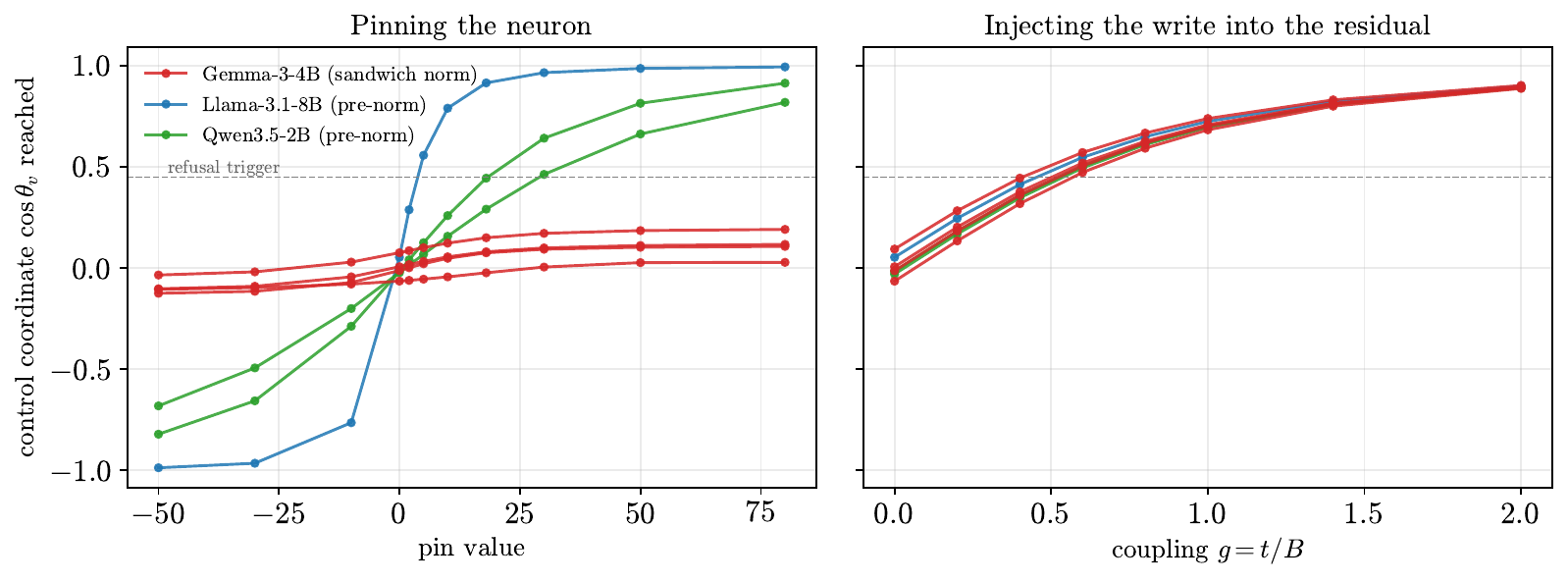}
\caption{A single neuron's reach under pinning versus residual injection, as the control coordinate $\cosv$ actually reached (mean over generic prompts). \textbf{Left, pinning the neuron's own activation:} on the pre-norm Llama-3.1-8B and Qwen3.5-2B the neuron drives $\cosv$ across the full range to $\pm0.99$, past the refusal trigger (dashed, $\cosv{=}0.45$); on Gemma-3-4B all four refusal neurons saturate at $\cosv\le0.19$ and never approach the trigger, because Gemma's post-FFN RMSNorm renormalizes the FFN output and caps the whole block's contribution to a small fraction of the residual norm. \textbf{Right, injecting the write directly into the residual:} the coordinate follows the universal drive on every architecture, reaching $\cosv\approx0.9$, so the cap is specific to what a \emph{neuron} can do, not to the coordinate. Pinning and injection therefore coincide for pre-norm models but diverge on Gemma. \emph{The vertical axis is the coordinate reached, not coherence}: injection drives Gemma's $\cosv$ high but the output collapses by $\cosv\approx0.3$ (the $m^\ast{=}0.31$ ceiling), so neither pinning (which cannot reach the trigger) nor injection (which collapses before it) coherently steers a Gemma refusal neuron.}
\label{fig:pinvsinject}
\end{figure}

Gemma's low ceiling is not the only reason its single-neuron windows are closed. The universal drive (Section~\ref{sec:law}) is measured by injecting the write into the residual, so it establishes that the \emph{coordinate} is reachable, not that a single neuron can reach it. We separate the two by sweeping each neuron under both interventions and reading the $\cosv$ actually reached (Figure~\ref{fig:pinvsinject}). On the pre-norm Llama and Qwen families pinning and injection coincide: the neuron drives $\cosv$ across the full range, past the refusal trigger ($\cosv\approx0.45$) to $\cosv\approx0.9$. On Gemma-3 all four audited refusal neurons \emph{saturate} at $\cosv\le0.19$ under pinning at any value up to $\pm80$, because the post-FFN RMSNorm renormalizes the FFN output and caps the whole block's contribution to a small fraction of the residual norm; direct injection still reaches $\cosv\approx0.9$. \textbf{A single Gemma neuron is therefore structurally unable to traverse its own coordinate}: the window is closed by the drive, not by the ceiling or the trigger. This scopes the pinning--injection equivalence of Section~\ref{sec:setup} to pre-norm models and is the third closed-window mechanism of Section~\ref{sec:prospective}; as a corollary, the placement of residual normalization is a structural lever on single-neuron controllability.

\subsection{Cross-architecture granularity map}
Table~\ref{tab:granularity} summarizes where coherent single-neuron control was found, by behavior and
architecture. Coherent safety bypass is open on the Qwen and Llama families and closed on Gemma;
mode-switch control (routing, operator) appears where the corresponding neuron exists; arithmetic K=1
control has a layer-count threshold near $32$--$36$ layers.

\begin{table}[h]
\centering
\caption{Cross-architecture single-neuron control. ``safety'' = coherent refusal bypass; ``mode-switch''
= routing or operator flip; window status is the dominant verdict for that model.}
\label{tab:granularity}
\begin{tabular}{llll}
\toprule
model & safety K=1 & mode-switch K=1 & window status \\
\midrule
Qwen2.5-3B & yes (\neuron{21}{2722}, \neuron{12}{4593}, +5) & --- & open, soft \\
Qwen3.5-2B & yes (\neuron{18}{1883}) & operator (\neuron{16}{2493}) & open \\
Qwen3-4B & --- (deep) & routing (\neuron{25}{347}) & open (routing) \\
Llama-3.1-8B & yes (\neuron{27}{13025}; \neuron{11}{4258}$^\dagger$) & operator (\neuron{19}{13312}) & open \\
Qwen3.5-4B & weak & --- & narrow \\
Gemma-3-4B & no & --- & \textbf{closed} \\
\bottomrule
\end{tabular}
\end{table}
$^\dagger$\neuron{11}{4258} is open only under budget-calibrated dosing (Section~\ref{sec:unify}).

%% file: sections/appendix_prospective.tex
\section{Prospective sweep, the collapse cliff, and trigger predictors}
\label{app:prosp}

\subsection{Prospective control-window sweep (full set)}
Table~\ref{tab:r1full} is the complete fifteen-neuron predict-then-dose test summarized in
Section~\ref{sec:prospective}. Predictions were committed before dosing. The committed verdict held on
eleven; all fifteen are consistent with the law. The four non-hits are an over-strict threshold call
(\neuron{24}{2598}), two decoupled closes at extreme depth (\neuron{32}{258}, \neuron{34}{2637}), and a
recovered dosing artifact (\neuron{31}{3309}); none violate $\cstrig<\cscoll\Leftrightarrow$ control. One
open neuron, \neuron{29}{2745} at relative depth $0.81$, was committed with an observed ceiling ($0.57$) and
trigger ($0.67$) that inverted. A finer, branch-resolved re-measurement resolves the inversion as a
wrong-sign measurement artifact: the original ceiling was set by an output-diversity collapse on the
\emph{non-controlling} ($+\vv$) branch, where no behavior flip occurs, whereas on the controlling ($-\vv$)
branch the output stays coherent across the dose grid and the behavior flips at $|\cosv|\approx0.40$. The
controlling-branch ceiling therefore lies at the top of the grid ($\gtrsim0.85$), matching the predicted
$0.82$, so \neuron{29}{2745} is a clean open with trigger well below ceiling; its corrected ceiling error
($\approx0.03$) is smaller than the committed $0.25$. We retain the committed values in the table for
integrity and note the correction here: recomputing the committed table's point-estimate MAE with the
corrected \neuron{29}{2745} ceiling lowers it from $0.139$ to $0.125$ and removes the only
trigger-above-ceiling inversion; the main text reports the committed point estimate of $0.14$, with bootstrap CI $0.09$--$0.23$,
computed on the committed values. The correction does not affect the open-direction conclusion or the
verdict count. The lesson generalizes: a ceiling must be read on the controlling branch, since a
sign-blind diversity collapse on the inactive branch can masquerade as a low ceiling. A branch-resolved
audit of all fifteen neurons found the same risk pattern (a non-controlling-branch collapse without any
behavior flip) on ten, with \neuron{29}{2745} the only case where it materially changed a published
number; the decoupled closes (\neuron{32}{258}, \neuron{34}{2637}) are genuine low-ceiling outcomes on
the controlling branch, not sign artifacts.

\begin{table}[h]
\centering
\caption{Full prospective sweep, fifteen held-out safety neurons over five architectures, all predictions
committed before dosing. ``P''/``O'' are predicted/observed. A dash in $\cstrig$ marks a closed or
artifact outcome with no coherent trigger. Bold marks the law's informative edges: the two closed
mechanisms (geometric on Gemma, decoupled at extreme depth) and the recovered dosing artifact. Ceiling
MAE $=0.14$ (bootstrap $95\%$ CI $0.09$--$0.23$); bulk (depth $\le0.8$) MAE $\approx0.07$.}
\label{tab:r1full}
\resizebox{\linewidth}{!}{%
\begin{tabular}{llccccll}
\toprule
model & neuron & depth & $\cscoll$ P & $\cscoll$ O & $\cstrig$ P & $\cstrig$ O & verdict P\,$\to$\,O \\
\midrule
Qwen2.5-3B & \neuron{20}{1378} & 0.56 & 0.88 & 0.81 & 0.45 & 0.31 & open\,$\to$\,open \\
Qwen2.5-3B & \neuron{20}{6831} & 0.56 & 0.88 & 0.76 & 0.45 & 0.42 & open\,$\to$\,open \\
Qwen2.5-3B & \neuron{8}{397}   & 0.22 & 0.93 & 0.86 & 0.60 & 0.67 & open\,$\to$\,open \\
Qwen2.5-3B & \neuron{29}{2745} & 0.81 & 0.82 & 0.57 & 0.45 & 0.67 & open\,$\to$\,open \\
Qwen2.5-3B & \neuron{32}{258}  & 0.89 & 0.78 & 0.46 & 0.45 & --- & open\,$\to$\,\textbf{closed (dec.)} \\
Qwen3.5-2B & \neuron{14}{5903} & 0.58 & 0.82 & 0.87 & 0.45 & 0.43 & open\,$\to$\,open \\
Qwen3.5-2B & \neuron{17}{5987} & 0.71 & 0.82 & 0.83 & 0.45 & 0.43 & open\,$\to$\,open \\
Qwen3.5-2B & \neuron{19}{723}  & 0.79 & 0.80 & 0.74 & 0.45 & 0.67 & open\,$\to$\,open \\
Qwen3-4B   & \neuron{22}{1149} & 0.61 & 0.81 & 0.84 & 0.45 & 0.70 & open\,$\to$\,open \\
Qwen3-4B   & \neuron{28}{1206} & 0.78 & 0.79 & 0.84 & 0.45 & 0.22 & open\,$\to$\,open \\
Qwen3-4B   & \neuron{34}{2637} & 0.94 & 0.74 & 0.27 & 0.45 & --- & open\,$\to$\,\textbf{closed (dec.)} \\
Llama-3.1-8B & \neuron{24}{2598} & 0.75 & 0.82 & 0.64 & 0.45 & --- & open\,$\to$\,\textbf{closed (marg.)} \\
Llama-3.1-8B & \neuron{31}{3309} & 0.97 & --- & 0.43 & --- & --- & artifact\,$\to$\,\textbf{open (rec.)} \\
Gemma-3-4B & \neuron{24}{132}  & 0.71 & 0.30 & 0.22 & 0.45 & --- & closed\,$\to$\,\textbf{closed (geo.)} \\
Gemma-3-4B & \neuron{31}{6160} & 0.91 & 0.28 & 0.09 & 0.45 & --- & closed\,$\to$\,\textbf{closed (geo.)} \\
\bottomrule
\end{tabular}}
\end{table}

\subsection{F4593 dose-response (the collapse cliff)}
Table~\ref{tab:cliff} reports the dose-response of the safety neuron \neuron{12}{4593}. Bypass is zero
below the predicted cliff $m_{\mathrm{cliff}}\approx9.7$, turns on between $|t|=10$ and $18$, and is
sign-asymmetric (the $+$ and $-$ directions carry the curvature and gradient channels respectively),
consistent with the dose-response being a finite-amplitude effect rather than a linear one.

\begin{table}[h]
\centering
\caption{Dose-response of \neuron{12}{4593} (safety, Qwen2.5-3B), bypass out of 24 prompts, coherence-gated bypass judge.
Cliff at the predicted $m_{\mathrm{cliff}}=\sqrt{2L_0/|\beta_H|}\approx9.7$, where $L_0$ is the baseline
margin to the decision boundary and $\beta_H$ the curvature of the dose-response (the second-order channel of
Section~\ref{sec:mechanism}; Appendix~\ref{app:trigger}, Equation~\eqref{eq:trigtheory}); sign asymmetry
$\mathrm{bp}(+50)/\mathrm{bp}(-50)=3.25$.}
\label{tab:cliff}
\begin{tabular}{lccc}
\toprule
dose $|t|$ & $\le 10$ & $+50$ & $-50$ \\
\midrule
bypass (/24) & 0 & \textbf{13} & 4 \\
\bottomrule
\end{tabular}
\end{table}

\subsection{Failed trigger predictors: subthreshold susceptibility}
\label{app:tables}
We attempted to predict reach from subthreshold rollout susceptibility, using symmetric dose ladders ($|g|$
up to $0.45$, below the control regime), sign-aware one-sided derivatives, and a path-integrated reach
$R_v$, with a refusal-projection compliance readout at increasing rollout length. After adding
\emph{layer-matched} random controls the apparent separation vanished: mean $R_v$ was $0.16$ for
controllers versus $0.13$ for layer-matched randoms (Qwen2.5-3B; $0.13$ versus $0.12$ on Llama), a random
neuron out-reached three of four controllers, and the subthreshold dose-response was flat. The earlier
separation against early-layer high-leverage neurons was a depth confound. The subthreshold readouts we
tested therefore do not predict the trigger; this negative result is consistent with a finite-amplitude
basin crossing (bypass appears only past the cliff of Table~\ref{tab:cliff}). We did not exhaust all
readouts---a logit-gap readout, a trained probe, and teacher-forced rollouts remain---so we scope the claim
to those tested. The reliable predictor remains the window relation between the trigger and the collapse
ceiling.

%% file: sections/appendix_surface.tex
\section{Control surfaces, negative behaviors, and the control taxonomy}
\label{app:negatives}

\subsection{Dose-by-rollout surfaces}
\label{app:surfaces}
\begin{figure}[t]
\centering
\includegraphics[width=0.86\linewidth]{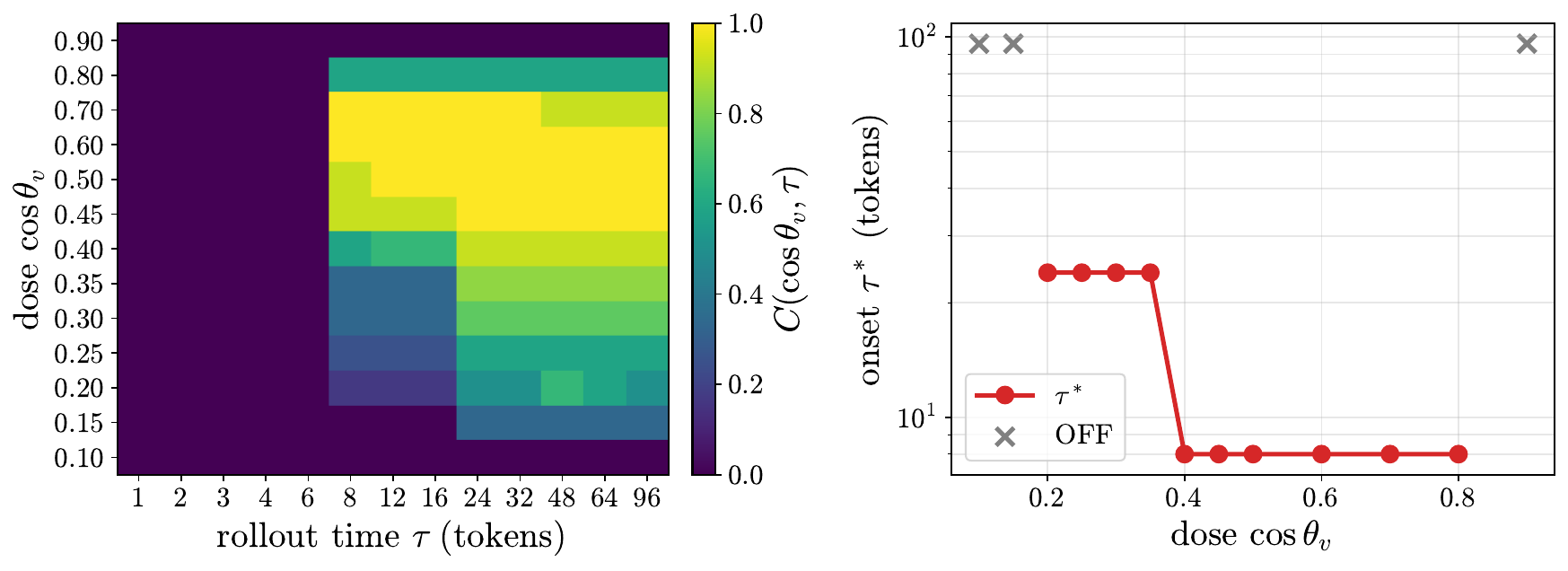}\\[1.5ex]
\includegraphics[width=0.86\linewidth]{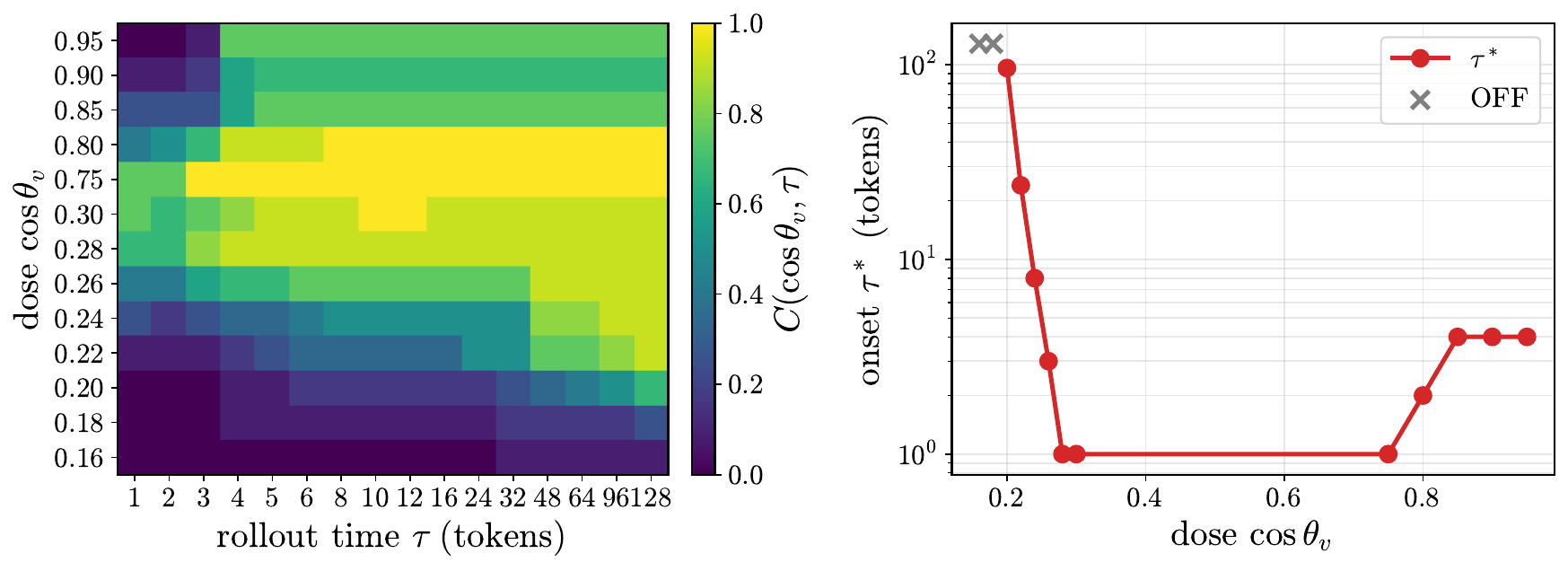}
\caption{Control surfaces. Top: operator (\neuron{19}{13312}) heatmap $C(\cosv,\tau)$ with a hard onset
floor at $\tau=8$ and total collapse at $\cosv=0.90$. Bottom: language (\neuron{25}{347}) boundary zoom,
token-local in the window with a diverging onset at the trigger edge. Each row pairs the heatmap (left) with
the onset curve $\tau^\ast(\cosv)$ (right).}
\label{fig:surfaces}
\end{figure}

Figure~\ref{fig:surfaces} shows the two exemplar control surfaces $C_i(\cosv,\tau)$ of
Section~\ref{sec:surface} as heatmaps with their onset curves. The operator surface (\neuron{19}{13312})
has a hard vertical floor at $\tau=8$: control is zero for $\tau\le6$ at every dose, switches on at
$\tau\ge8$ inside the window, and collapses entirely at $\cosv=0.90$. The language surface (\neuron{25}{347},
boundary zoom) is token-local in the window and shows the diverging onset latency at the lower edge
($\tau^\ast=96\to1$ over $\cosv=0.20\to0.28$). Onset times carry bootstrap $95\%$ intervals over the twelve
prompts, tight in the window (operator $[8,8]$) and wide at the boundary as expected near a continuous
transition, the early-warning signature of such transitions \citep{scheffer2009earlywarning}. The rising-edge
fit $\tau^\ast\propto(\cosv-c_\ast)^{-z}$ is phenomenological and used only to
estimate the boundary $c_\ast\approx0.15$; the exponent is not interpreted as universal.

\subsection{A behavior with no single-neuron gate: format routing}
\label{app:format}
Not every behavior we tried yielded a single-neuron controller. We searched for a benign format-routing gate
that turns a prose answer into JSON, scored by a deterministic detector requiring parseable JSON structure
rather than a mere opening brace. A brute-force causal scan of $1{,}200$ neurons across eight Llama layers found
none (window count $0/1{,}200$, baseline JSON rate $0$). A positive control confirmed the detector and the
generation path: prompting the model to answer in JSON produced parseable JSON on $6/6$ prompts. Because
JSON onset is token-local, we then ran a zero-generation first-token-logit probe over four full layers
($57{,}344$ neurons), scoring each neuron by the injected increase in the opening-structure logit relative
to its competitors, and verified the top twenty-four per layer by generation. The probe surfaced many
brace-pushing neurons, with first-token margins up to about nineteen and one neuron opening with a brace on
half its prompts, but none produced valid JSON on a majority of prompts (best $1/4$). Two readings follow.
First, the probe behaves as the onset axis predicts: a first-token score recovers token-local brace onset,
exactly the regime where it should work, and it correctly fails to confirm those candidates once parseable
JSON is required, the same necessary-but-not-sufficient pattern as the curvature screen of
Section~\ref{sec:gradient}. Second, we found no strict-onset prose-to-JSON single-neuron gate under two
matched probes, both with positive controls. This indicates that such gates are not common, or are not
captured by a strict first-token JSON criterion, but it does not rule out distributed, later-onset, or
prompt-specific format controllers. A second benign behavior points the same way: sort-direction routing,
where the prompt asks for ascending order and the target is a flip to descending, yielded no single-neuron
gate under a matched short-rollout generation scan of $1{,}200$ neurons (baseline $0$), with the onset-matched
probe ruling out a probe mismatch. The best descending rate was only $0.17$, and inspecting the generations
traced even that to a prompt artifact rather than control: it was exactly the two of twelve lists whose first
element is the list maximum, which flip to a descending run under any perturbation and do so across many
different neurons, so the neuron-specific descending rate is near zero. Single-neuron control is thus sparse and behavior-specific: of four behaviors we probed, two have a
gate (operator, language routing) and two do not under matched probes (format, sort). The dose-rollout
surface characterizes the behaviors that have a gate rather than asserting that all behaviors do.

\subsection{A control taxonomy}
\label{app:taxonomy}
The two positive controllers and the two negatives, together with the anchor-controlled follow-up, fall into
the classes of Table~\ref{tab:taxonomy}. The anchor experiment varied whether the input list to a
smallest-to-largest sort begins with its maximum element. Descending output appeared under perturbation
almost only when the first element was the maximum (rate $0.17$ versus $0.03$ for non-anchored lists, $0.01$
when the maximum was in the second position, and $0$ when the first element was the minimum; baseline $0$ in
all conditions). No neuron produced descending output on non-anchored lists ($0$ of $12$), so there is no
sort gate. The dominant effect of perturbation was disruption, not reversal: most non-ascending output was
scrambled rather than descending, with the disrupted fraction reaching $0.62$ on non-anchored lists and
$0.58$ with the maximum in the second position. Perturbation thus degrades the sort rather than flipping its
direction, and the small descending channel is the anchored exception. That channel is a neuron-by-prompt
interaction: the anchor is necessary but not sufficient, since only some neurons amplify it (per-neuron
amplification index $0$ to $0.83$, including one random neuron at $0.25$). We call this prompt-affordance amplification: perturbation amplifies a prompt-local heuristic in some neurons, but the effect is prompt-conditioned and neuron-promiscuous rather than a sparse, prompt-invariant gate, sitting between a sparse controller and no effect. This motivates two discovery guardrails: audit the positive examples, and separate target reversal from task disruption.

A controller-versus-affordance check makes the contrast direct. We measured the target-flip rate for each true controller and seven budget-matched random neurons across prompt-affordance conditions: for the operator, a few-shot cueing addition (anti-affordance), a neutral query, and one cueing multiplication; for the language gate, an explicit answer-in-English instruction (anti-affordance), a plain English question, and an English question on a Chinese topic. Both controllers flipped their target under perturbation across all conditions, including the anti-affordance ones (rate $1.0$), with one detector-format caveat on the operator's multiplication-cued few-shot. There the strict product-not-sum score fell, but inspecting the generations the gate still held the model in multiplication mode on all twelve prompts, with no reversion to addition; the multiplication few-shot induced verbose explanations and echoes of the few-shot examples that the strict format does not score. This is a detector-format effect rather than a weakening of control, and the decisive anti-affordance and neutral conditions are $1.0$. The language gate flipped to Chinese even when the prompt demanded English. No random neuron flipped in any condition ($0.0$), and no condition produced the target at baseline ($0.0$), so neither behavior has a self-sufficient prompt affordance. Affordance-invariance thus separates a sparse controller, neuron-specific and prompt-general, from prompt-affordance amplification, which needs the affordance and is neuron-promiscuous. Unlike the sort case, budget-matched random perturbations in the operator and language tasks generally preserved the default behavior, coherent addition or fluent English, rather than producing target flips, so random perturbation here is closer to no effect than to disruption.

\begin{table}[t]
\centering
\small
\caption{A working taxonomy of the observed single-neuron control and failure modes, one row per behavior we
probed. It is an empirical map of the cases we measured, not a claim that these are the only modes.}
\label{tab:taxonomy}
\begin{tabular}{p{0.24\linewidth} p{0.16\linewidth} p{0.18\linewidth} p{0.28\linewidth}}
\toprule
mode & neuron-specific & prompt-specific & example \\
\midrule
sparse controller & high & low to moderate & operator \neuron{19}{13312}, language \neuron{25}{347} \\
prompt-affordance amplification & partial / promiscuous & high (needs the affordance) & sort, max-first lists \\
distributed / no gate found & none found & varies & format routing (prose to JSON) \\
content-risk gate & medium to high & high domain interaction & safety \neuron{11}{4258}, \neuron{10}{4789} \\
\bottomrule
\end{tabular}
\end{table}

%% file: sections/appendix_safety.tex
\section{Content-harm audit and the alignment comparison}
\label{app:safetygrp}

\subsection{Content-harm audit, row level}
\label{app:content}
This expands the content-harm cohort of Section~\ref{sec:typed}. On Llama-3.1-8B six single-neuron refusal
pivots are window-open and audited by all three judges; at their peak over rollout depth three reach
strict-actionable content---\neuron{11}{4258} and \neuron{10}{4789} at $0.88$ and \neuron{10}{1084} at
$0.75$---while \neuron{15}{9635} reaches only $0.17$ and \neuron{16}{6950} and \neuron{17}{13532} bypass
refusal coherently with zero actionable content on every dose and horizon. The newly identified
\neuron{10}{4789} matches the published refusal neuron \neuron{11}{4258} on the strict-actionable peak under
the identical audit, and attains that peak at shorter rollout horizons (its actionable rate plateaus by
$\tau\approx96$ against $\tau\approx128$) and at a milder dose. Two earlier candidates do not survive the
strict audit: \neuron{27}{13025}, which a coarse two-way read had called a content window, resolves under a
register-aware re-judge into coherent non-refusal and unsafe-topic setup with zero actionable compliance,
and a sign-and-dose sweep of \neuron{24}{2598} opens only a weak window. Where harm is genuine it
concentrates on the cyber and fraud prompts that single-neuron steering elicits most readily, on
\neuron{11}{4258}'s negative sign. On Qwen the strongest pivots (e.g.\ \neuron{21}{2722}, Qwen2.5-3B;
\neuron{28}{1206}, Qwen3-4B; \neuron{20}{1378}, Qwen2.5-3B; \neuron{14}{5903}, Qwen3.5-2B) produce coherent
non-refusal or first-token refusal-template flips but no actionable harm once both signs are swept---unsafe
-topic setup and semantic collapse that an automated content classifier over-flags, with zero genuine
compliance on audit. The genuine actionable population is therefore the Llama subset; the Qwen rows are
proxy artifacts rather than weaker instances of the same effect.

The prefix audit also times the onset of completed harm (Section~\ref{sec:contract}). Across the Llama
pivots with validated content windows, prefixes typically reach harmful setup by the first $32$ tokens,
while strict actionable compliance appears later, plateauing by $\tau\approx96$ to $128$ and then holding
flat to $\tau256$ (a verified prefix-superset, so the actionable rate is a cumulative prefix-audit); the
grid is coarse, so the true onset lies between $64$ and $128$. A recall-oriented classifier that fires on
the early setup therefore over-counts and mis-times completed harm: a first-hit detector reads harm as
present within $16$ tokens, where the strict content judge finds only setup. The split holds across these
pivots, so the early-setup, late-actionability pattern is not specific to a single neuron.

\paragraph{The two-axis robustness diagnostic.} Table~\ref{tab:rwindow} reports the per-model two-axis
diagnostic introduced in Section~\ref{sec:unify}: the coherent-bypass margin and the strict-actionable
potency, the defensive deliverable a model owner can compute and drive toward zero.

\begin{table}[h]
\centering
\small
\caption{Two-axis refusal diagnostic, reported per model. The coherent-bypass margin $\mathcal{R}_{\mathrm{bypass}}=\max_i(\cos\theta^{\mathrm{coll}}_{v,i}-\cos\theta^{\mathrm{trig}}_{v,i})_+$ is the widest coherent-bypass window width over the audited refusal neurons for that model; the strict-actionable peak is the highest content-confirmed actionable rate over the same audit (peak over $\tau$); $\tau_{\mathrm{actionable}}$ is the first rollout horizon at which that peak appears. The table is a calibration diagnostic, not a leaderboard: a model can open a coherent bypass window with zero strict-actionable reach. Rows are reported per model; $\mathcal{R}_{\mathrm{bypass}}$ and the strict-actionable peak are maxima over the audited refusal neurons for that model. $^\dagger$\neuron{28}{1206}'s committed trigger is the least certain entry; a branch-resolved re-audit brackets its bypass margin at $0.49$--$0.60$. Strict-actionable peaks are over $n=24$ prompts (Wilson $95\%$ half-width $\le0.19$); $\mathcal{R}_{\mathrm{bypass}}$ is resolved to one dose-grid bin.}
\label{tab:rwindow}
\begin{tabular}{lcccl}
\toprule
model & $\mathcal{R}_{\mathrm{bypass}}$ & strict-actionable peak & $\tau_{\mathrm{actionable}}$ & typed outcome \\
\midrule
Llama-3.1-8B & \textbf{0.65} & 0.88 & $\approx96$--$128$ & genuine-actionable + bypass-only \\
Qwen3-4B$^\dagger$ & $0.49$--$0.60^\dagger$ & 0.00 & --- & proxy / first-token flip \\
Qwen2.5-3B & 0.50 & 0.00 & --- & proxy / setup \\
Qwen3.5-2B & 0.44 & 0.00 & --- & proxy / setup \\
Gemma-3-4B & 0 & 0.00 & --- & closed (geometric) \\
\bottomrule
\end{tabular}
\end{table}

\subsection{Alignment appears to install the trigger}
\label{app:alignment}
The window's two edges have different origins. Section~\ref{sec:prospective} traced the ceiling to
architecture geometry; the trigger, the under-determined edge, appears to be installed by alignment training.
A base-to-instruct comparison of seven safety pivots suggests the discrimination structure pre-exists
alignment and that training converts it into a behavioral gate. The write directions are largely preserved,
the Qwen pivots at weight cosine above $0.998$ and the Llama pivot \neuron{11}{4258} instead rewriting its
direction, while the harmful-versus-benign discriminative signal the late circuit reads is amplified by up to
three orders of magnitude, an amplification absent on the arithmetic pivot \neuron{19}{13312} and so
safety-specific. In window terms, base models have a readable direction but no reachable trigger, and
alignment installs the trigger by amplifying the feature the late circuit reads; single-neuron bypass then
works by forging that amplified signal, saturating the feature so the safety circuit reads ``harmless'' where
harm should register. This is the base-versus-instruct form of readability without causal use, corroborating
the concurrent report that alignment ``transforms function, not structure'' \citep{herring2026cna}; we read
it as an evidence-backed interpretation of where the trigger comes from, not an established mechanism.

Table~\ref{tab:alignment} reports the comparison. At the weight level the write columns of seven safety
pivots are nearly unchanged from base to instruct, the Qwen pivots almost exactly; the Llama pivot
\neuron{11}{4258} is the exception and follows the alternative recipe of rewriting its direction. At the
activation level the same neurons' discriminative gap between harmful and benign prompts is amplified by up
to three orders of magnitude, and the amplification is safety-specific: the arithmetic pivot
\neuron{19}{13312} shows none on safety prompts. In base models, steering these neurons shifts content
without changing the decision; in instruct models, the same neurons gate the behavior. Two family recipes
appear: Llama rewrites and amplifies, Qwen amplifies while preserving columns.

\begin{table}[h]
\centering
\small
\caption{Alignment appears to install the trigger by preserving the write direction and amplifying the
signal it carries. Weight cosine is base to instruct; signal amplification is the harmful-versus-benign
discriminative gap, base to instruct. Two recipes appear (Qwen preserves and amplifies; Llama
\neuron{11}{4258} rewrites), and the arithmetic pivot is the safety-specificity control.}
\label{tab:alignment}
\begin{tabular}{llcc}
\toprule
neuron & model & weight cosine (base\,$\to$\,instruct) & signal amplification \\
\midrule
\neuron{12}{4593} & Qwen2.5-3B & 1.0000 & $4\times$ \\
\neuron{20}{1378} & Qwen2.5-3B & $>0.998$ & $1650\times$ \\
\neuron{11}{4258} & Llama-3.1-8B & rewritten (top $0.15\%$ in layer) & --- \\
\neuron{19}{13312} (arithmetic) & Llama-3.1-8B & --- & none on safety \\
\bottomrule
\end{tabular}
\end{table}

%% file: sections/appendix_measurement.tex
\section{Measurement: coherence-gating, dose-calibration, and artifact corrections}
\label{app:measure}

This appendix gives the artifact analysis behind the measurement contract of Section~\ref{sec:contract}.
Behavior judges over-count control unless two corrections are made: gate the behavior on coherence, and
calibrate the dose in units of the budget. Both are required for every window measurement in the paper, and
both correct errors in prior reports.

\textbf{Gate behavior on coherence.} A behavior detector applied to degenerate text reports spurious success. At a collapse-edge dose, a refusal neuron emits incoherent output that a refusal-string detector scores as ``not refusing,'' and a language router emits repetitive Chinese characters that a character detector scores as a successful switch. We therefore score a flip only when the output is coherent, using the distinct-bigram ratio \citep{li2016diversity} as a threshold-light gate. The same gate, applied to a fixed-magnitude language-injection result, would have caught a reported success that was in fact degenerate repetition.

\textbf{Use the distinct-bigram ratio, not entropy, and match the gate to the script.} Degeneracy is non-monotonic in next-token entropy \citep{holtzman2020degeneration}. Collapse takes two forms: low-entropy repetition, when the pinned direction maps to a coherent token (typical of late, readout-aligned neurons), and high-entropy scramble, when it maps off the token manifold (typical of early neurons). Coherent generation sits between these extremes, so no single entropy threshold separates coherent from collapsed. The distinct-bigram ratio is monotone in degeneracy because greedy decoding of either collapse mode emits low-diversity sequences. The gate is not universal across scripts, however: computed over whitespace-delimited words it is identically $1.0$ on unspaced output such as Chinese, so for CJK and other unspaced scripts we gate on character diversity, the fraction of distinct characters in the generation, which is script-agnostic. And a length gate alone is never a coherence gate: it false-fires on short correct answers and can even invert a capability metric by rewarding long repetition over a clean short reply.

\textbf{Distinguish a closed window from a failed measurement.} A behavior that never registers can mean the window is closed (the real result) or that the detector could not read coherent output (an artifact). We separate them with the count of parseable generations and the coherence ratio: a closed window shows the behavior staying flat while coherence is intact up to collapse, whereas a failed measurement shows no parseable output while coherence is high. One operator run that initially read as null was, by this check, a parser failure (one of twelve prompts parsed at baseline) and was corrected by a longer rollout.

\textbf{Branch-resolve the window.} The control coordinate is signed, and the two edges of the window must be measured on the same controlling sign of $\vv$. The danger is asymmetric collapse: on a sign-specific controller the non-controlling branch often collapses at high dose \emph{without any behavior flip}, so a sign-blind ceiling search reports that collapse as the ceiling and can manufacture a closed verdict, or an apparent trigger-above-ceiling inversion, on a neuron whose controlling branch is cleanly open. The one such inversion in our prospective sweep resolved exactly this way: the committed ceiling of \neuron{29}{2745} was the non-controlling-branch collapse, and the branch-resolved re-measurement restores a clean open window matching the predicted ceiling (Appendix~\ref{app:prosp}). The same discipline requires storing the baseline alignment $\cosv(0)$, not only the driven coordinate, since a non-orthogonal baseline shifts where each branch starts (Section~\ref{sec:law}). The discipline extends to \emph{trigger} fits, not only ceilings: a local margin expansion fit on a sign-averaged grid mixes the controlling and non-controlling branches and manufactures an apparent silent-gate over-prediction of the trigger, which a branch-resolved fit removes (Appendix~\ref{app:trigger}). Branch resolution is thus the same correction at both edges of the window.

\textbf{Rule out generic disruption.} A single write could in principle flip a behavior by degrading the computation broadly rather than steering the specific behavior; three controls separate the two. The response is \emph{behavior-specific}: the same kind of injection switches a router to a particular language, flips an operator from addition to multiplication, and turns refusal into compliance, each read by a distinct detector, whereas generic degradation tracks no specific behavior. It is often \emph{sign-specific}: many controllers open a window on only one sign of $\vv$, such as the recovered \neuron{31}{3309} on $-\vv$, which a sign-blind content disruption cannot produce; a minority instead open coherent windows on \emph{both} signs (for example \neuron{20}{1378} and \neuron{20}{6831} in Qwen2.5-3B), which we note at the behavior level only: a per-sign content audit with a positive control found no support for two-way \emph{content} routing (any residual signal was the same category on both signs), though at the long harvest required for content the generations degrade, so we leave content-level branch semantics to prefix-audited future work. And the coherence gate already removes flips that coincide with degraded text. The closed-window results carry a matching \emph{positive control}: open windows on other architectures show that the detector and dose range do register control when it exists, so Gemma's null is a real closed window, marked by coherence actively falling to zero rather than by silent flatness, and not a failure to measure.

\textbf{Not-refusal is not harm: content needs a second gate.} Coherence-gated non-refusal still over-counts genuine harm. A neuron that flips the refusal template can produce coherent text that is not actually harmful-actionable, and at collapse-edge doses an automated safety classifier flags incoherent fragments as unsafe. We therefore separate two questions: whether the model stops refusing (the behavior we control) and whether it produces genuinely harmful content (a content judgment). The content judge itself must be chosen with care: an aligned instruct model refuses the meta-task and can never emit a positive label, so we use a purpose-built safety classifier (high recall but low precision, over-flagging collapsed text) as a screen and then gate it.

Critically, the lexical coherence gate is necessary but not sufficient, because collapse has two forms and the distinct-bigram gate catches only one. \emph{Lexical collapse} (repetition, scramble, malformed loops) drives the distinct-bigram ratio down and is caught; \emph{semantic collapse} keeps the ratio high while the text drifts off task into template, entity-extraction, or boilerplate filler, and slips the lexical gate. We see the latter at large dose on Qwen, where outputs stay lexically coherent but become off-task. We therefore judge in four steps: whether the reply is coherent at the surface, whether it is on-task for the request, whether an on-task reply complies or refuses, and whether a complying reply is specific and actionable. Genuine harm is the intersection of all four (lexical coherence, task-faithfulness, compliance, and actionable content), so that fluent off-task text is scored as neither a safe refusal nor a harmful compliance. An automated controller search scored by non-refusal and lexical coherence alone inherits the same gap, admitting high-dose semantic collapse as apparent control. A finer prefix audit, scoring the same generation truncated at a fine token grid, further separates a generation that has merely entered a harmful setup from one that is already actionable, and shows that actionable compliance commits late in the rollout (Appendix~\ref{app:content}). The prefix audit also guards against the converse error: because a generation can deliver content and \emph{then} degrade, gating a content event on full-reply coherence deletes it, and in a per-sign audit such an AND-gate zeroed out even a known-harmful positive control whose raw classifier rate was $0.94$. Content must therefore be scored at the event level on prefixes, with tail coherence reported separately, while the behavioral window verdict keeps its coherence requirement at the behavior's own horizon. Across our content-judged neurons, non-refusal over-counts harm severely, and genuine harmful compliance is dose-windowed and model-specific (Section~\ref{sec:typed}).

\textbf{Register-aware judging.} Heuristic refusal judges over-count bypass through at least four failure modes: repetition loops, curly-apostrophe mismatches, second-person or crisis-reframed refusals, and soft academic deflections. A three-way judge that separates compliance from template and reframed refusal is required. The published refusal neuron \neuron{11}{4258}, scored as compliant by a two-way judge, is reclassified as degraded once coherence is gated, consistent with the capability cost \citet{kazemi2026} report for the constant pin and with the dose analysis of Section~\ref{sec:unify}. We report every behavior count with its denominator and the judge that scored it, and we treat the correction of an earlier over-claim as evidence that the framework is falsifiable rather than as a defect.

%% file: sections/appendix_examples.tex
\section{Control windows, the neuron census, and selected examples}
\label{app:gallery}

\subsection{All measured control windows (full set)}
\label{app:windows-full}
\begin{figure}[h]
\centering
\includegraphics[width=\linewidth]{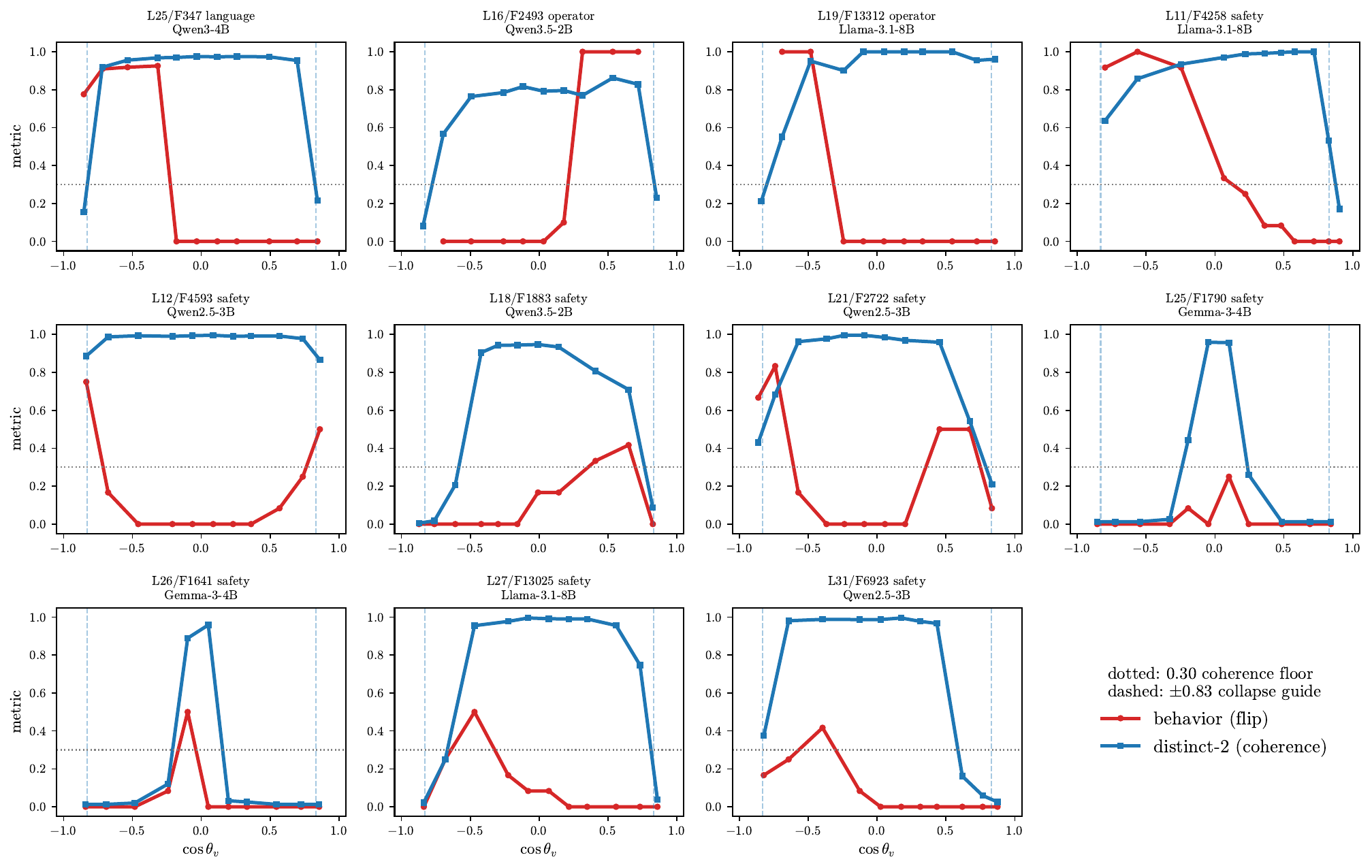}
\caption{Control windows for eleven controllers across three behavior classes (full set; the representative
subset is Figure~\ref{fig:windows}). Each panel plots the behavior metric (red) and coherence, the
distinct-bigram ratio (blue), against the control coordinate $\cosv$; the dotted line is the $0.30$
coherence floor and the dashed lines the $\pm0.83$ collapse guides.}
\label{fig:windows-full}
\end{figure}

Figure~\ref{fig:windows-full} is the full eleven-controller version of the representative
Figure~\ref{fig:windows}. Mode switches (routing, operator) trigger sharply near $\cosv\approx0.3$ with wide
coherent windows; coarse safety decisions trigger softly near $0.45$ with narrower windows; the two Gemma
neurons (\neuron{25}{1790}, \neuron{26}{1641}) collapse before any trigger, the closed-window case.

\subsection{Selected examples}
\label{app:examples}
This gallery collects the single-neuron cases across the gate types of
Table~\ref{tab:taxonomy}, deliberately broader than safety: safety is one gate type among several, not the
focus. Table~\ref{tab:census} is the full census of featured neurons with their model, type, and one example
each; the paragraphs that follow add a short discovery story for each. The fifteen further held-out validation
neurons are in Table~\ref{tab:r1full}, and examples for safety neurons are content-free.

\begin{table}[h]
\centering
\small
\caption{Census of featured single-neuron cases, grouped by gate type, with model and one example each.
Safety examples are content-free; the row-level content audit is Appendix~\ref{app:content} and the fifteen
held-out validation neurons are Table~\ref{tab:r1full}. All neurons were identified in this work except the
published refusal neuron \neuron{11}{4258} \citep{kazemi2026} and the super weight \citep{yu2024superweight}.}
\label{tab:census}
\begin{tabular}{llp{0.21\linewidth}p{0.34\linewidth}}
\toprule
neuron & model & type & one example \\
\midrule
\neuron{25}{347} & Qwen3-4B & routing gate (mode switch) & ``What is a computer?'' $\to$ answer rendered in Chinese \\
\neuron{19}{13312} & Llama-3.1-8B & operator gate (mode switch) & ``$3+4=$'' $\to$ ``$3\times4=12$'' \\
\neuron{16}{2493} & Qwen3.5-2B & operator gate (mode switch) & addition prompt $\to$ product (operator flipped) \\
\midrule
\neuron{8}{3742} & Qwen3.5-2B & framing gate (task-frame) & arithmetic $\to$ ``cannot solve\ldots mathematically impossible'' template ($\cstrig\approx0.6$) \\
\neuron{8}{645} & Qwen3.5-2B & framing gate (task-frame) & arithmetic $\to$ logical-deduction template \\
\neuron{10}{4780} & Qwen3.5-2B & framing gate (task-frame) & arithmetic $\to$ question-rephrasing template ($-\vv$) \\
\neuron{16}{3146} & Llama-3.1-8B & framing gate (task-frame) & arithmetic $\to$ calculation/complexity template (pin $\ge30$) \\
\midrule
\neuron{11}{4258} & Llama-3.1-8B & safety gate, content-risk & refusal $\to$ coherent non-refusal, dose-corrected (content withheld) \\
\neuron{10}{4789} & Llama-3.1-8B & safety gate, content-risk (identified here) & refusal $\to$ actionable content, matches the published pivot under audit (content withheld) \\
\neuron{27}{13025} & Llama-3.1-8B & safety gate, coherent bypass only & refusal $\to$ non-refusal, no actionable content on register-aware audit (content withheld) \\
\neuron{18}{1883} & Qwen3.5-2B & safety gate, soft & refusal $\to$ non-refusal at a soft trigger \\
\neuron{12}{4593} & Qwen2.5-3B & safety gate, soft & refusal $\to$ non-refusal past a finite-amplitude cliff \\
\neuron{20}{1378} & Qwen2.5-3B & safety gate, soft & refusal $\to$ non-refusal (open, prospective) \\
\neuron{8}{397} & Qwen2.5-3B & safety gate, early-layer & non-refusal only at a raised trigger ($\cosv\approx0.67$) \\
\midrule
\neuron{21}{2722} & Qwen2.5-3B & safety, template flip (no genuine harm) & refusal template flips, content stays non-actionable \\
\neuron{24}{2598} & Llama-3.1-8B & safety, weak / marginal & weak window below the coherence gate \\
\midrule
\neuron{25}{1790} & Gemma-3-4B & closed window, geometric & every coherent dose stays refusal, then collapses \\
\neuron{26}{1641} & Gemma-3-4B & closed window, geometric & ceiling below any trigger \\
\neuron{24}{132} & Gemma-3-4B & closed window, geometric & predicted closed, confirmed \\
\neuron{32}{258} & Qwen2.5-3B & closed window, decoupled (deep) & no flip at any coherent dose, then collapse \\
\neuron{34}{2637} & Qwen3-4B & closed window, decoupled (deep) & no reachable trigger at depth $0.94$ \\
\midrule
\neuron{31}{3309} & Llama-3.1-8B & recovered artifact & fixed-pin ``destructive'' $\to$ calibrated $-\vv$ coherent bypass \\
\midrule
super weight & cross-architecture & scale-setter (not a controller) & ablation collapses the model; selects no behavior \\
(promiscuous) & Qwen / Llama & prompt-affordance amplification & max-first list $\to$ descending; neuron-promiscuous \\
(none found) & Llama-3.1-8B & no single-neuron gate & prose $\to$ JSON: brace only, not valid JSON ($0/57{,}344$) \\
\bottomrule
\end{tabular}
\end{table}

\paragraph{Language router (sparse controller, mode switch).} \emph{Type:} routing gate.
\emph{Neuron:} \neuron{25}{347}. \emph{Model:} Qwen3-4B. Surfaced by perturbation probing as a low-gradient
neuron that the FFN-to-skip diagnostic flagged. Under a calibrated dose it switches the generation language
from English to Chinese on $12/12$ prompts with a token-local onset ($\tau^\ast=1$). It is a clean
counterexample to gradient ranking: it sits in the bottom few percent of the behavior-token gradient yet
controls coherently, and it flips even when the prompt instructs the model to answer only in English.
\emph{Example.} Prompt (English): ``What is a computer?'' $\to$ the dosed response is rendered entirely in
Chinese (glossed: ``a computer is a tool that performs tasks through electronic devices\ldots''), whereas the
baseline answers in English.

\paragraph{Arithmetic operator gate (sparse controller, mode switch).} \emph{Type:} operator gate
($+\!\to\!\times$). \emph{Neurons:} \neuron{19}{13312}, \neuron{16}{2493}. \emph{Models:} Llama-3.1-8B,
Qwen3.5-2B. Found from the same two-forward-pass primitive after both gradient and subspace attribution
missed it. On symbolic arithmetic $a+b{=}$ it switches addition to multiplication under a budget-calibrated
dose and is scored by exact match, so it is the cleanest judge-free instance of the law; its flip is absent
from the next-token distribution and develops over a hard onset floor of about eight tokens.
\emph{Example.} Prompt: ``$3 + 4 =$'' $\to$ ``$3 \times 4 = 12$'': the gate rewrites the operator and returns
the product, scored by exact match against the sum.

\begin{figure}[h]
\centering
\includegraphics[width=0.62\linewidth]{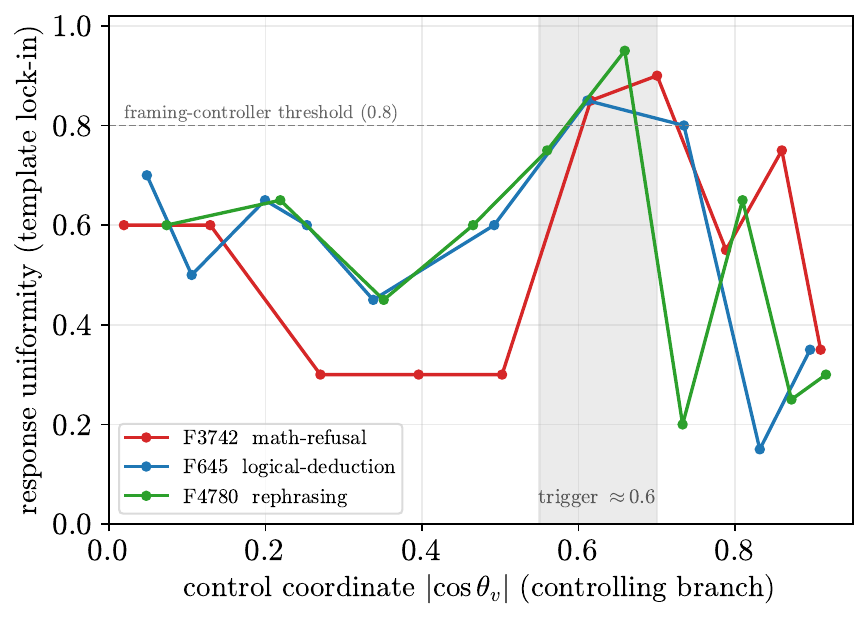}
\caption{Framing controllers obey the control window. Response uniformity (template lock-in on benign
arithmetic prompts, a judge-free oracle) against the control coordinate $|\cosv|$ on each neuron's controlling
sign, for the three confirmed Qwen3.5-2B framing controllers. The template locks in, crossing the $0.8$
controller bar, across a trigger band near $|\cosv|\approx0.6$ (shaded), higher than the mode-switch trigger
($\approx0.3$) and refusal ($\approx0.45$); coherence does not collapse in range, so the window is open.
\neuron{8}{3742} is the math-refusal controller (``mathematically impossible'' on a simple sum).}
\label{fig:framingwindow}
\end{figure}

\paragraph{Task-framing controller (sparse controller, response template).} \emph{Type:} framing gate.
\emph{Neurons:} \neuron{8}{3742}, \neuron{8}{645}, \neuron{10}{4780} (Qwen3.5-2B), \neuron{16}{3146}
(Llama-3.1-8B). A framing controller changes neither the content nor the operator; it forces the response into
a fixed \emph{template} (a logical-deduction preamble, a question-rephrasing, or a task-refusal frame), scored
judge-free by response uniformity over benign arithmetic prompts, an exact oracle like the operator's. The most
striking is \neuron{8}{3742}, which at a calibrated dose makes the model answer every simple sum with ``No, I
cannot solve this problem because it is mathematically impossible'' at $100\%$ uniformity. These controllers
obey the control window on each neuron's controlling sign with a behavior-class trigger $\cstrig\approx0.6$
(Figure~\ref{fig:framingwindow}), higher than the mode-switch $\approx0.3$ and the refusal $\approx0.45$, and they confirm the dose-calibration
correction on a new class: \neuron{16}{3146}'s framing verdict is non-monotonic in a fixed pin (inert at $+10$,
framing at $+30$ and $+80$) yet resolves cleanly on the budget coordinate. \emph{Example.} Prompt: ``What is
$46 + 84 = ?$'' $\to$ ``No, I cannot solve this problem because it is mathematically impossible\ldots'' (the
sum is $130$).

\begin{figure}[h]
\centering
\includegraphics[width=0.82\linewidth]{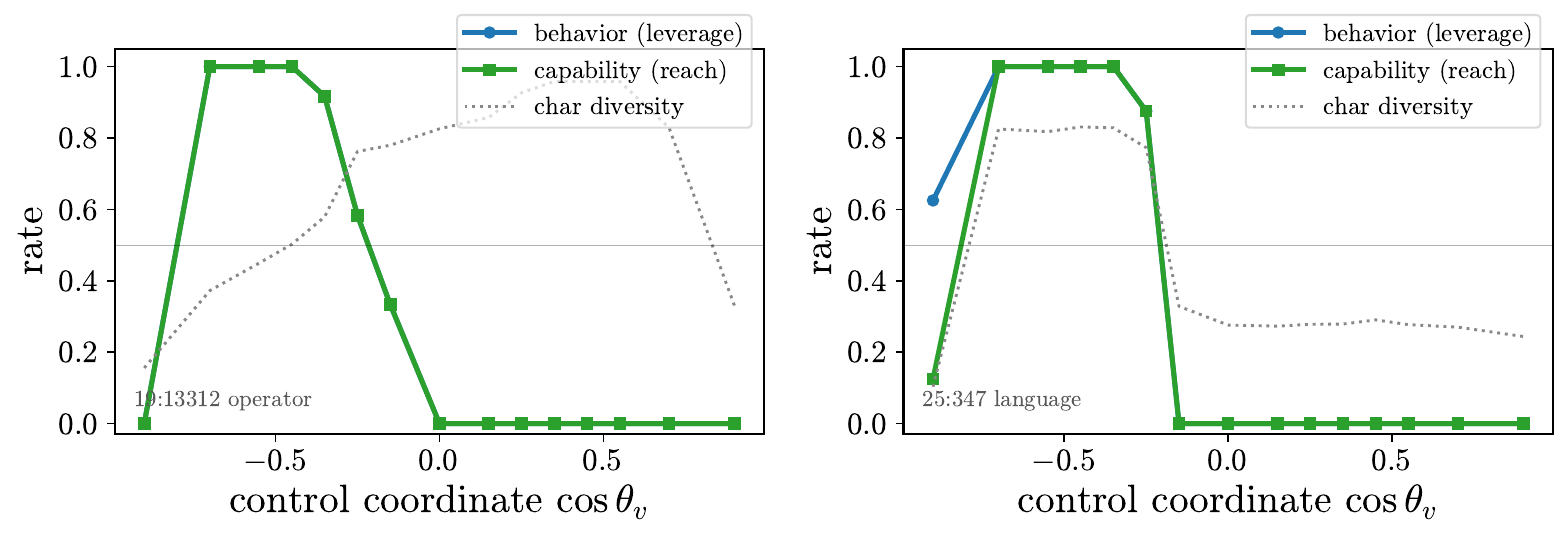}
\caption{Inside the window the dose preserves task-local capability. For the benign mode-switch neurons (operator $+\!\to\!\times$, \neuron{19}{13312}; language EN$\to$ZH, \neuron{25}{347}), the behavior rate (leverage), the capability rate (reach: on-target \emph{and} non-degenerate), and character diversity vs the control coordinate $\cosv$. Capability is high and flat from the trigger across the controlling ($-\vv$) window and is zero on the non-controlling branch. For the language router the curves separate at the ceiling (the dose still routes the language while the text collapses), making ``leverage is not reach'' quantitative; for the operator, whose detector requires a coherent correct product, behavior and capability coincide and fall together at the ceiling.}
\label{fig:capability}
\end{figure}

\paragraph{Capability inside the window (benign mode switches).} On the benign mode-switch tasks the control
window is where \emph{task-local capability} survives and is redirected, not merely where behavior flips.
Inside its window the operator gate \neuron{19}{13312} computes the correct \emph{product} ($3+4\!\to\!12$)
rather than merely stopping addition, and the language router \neuron{25}{347} emits fluent, on-topic
target-language text. Figure~\ref{fig:capability} plots the capability rate, the fraction of generations that
are on-target and non-degenerate, against $\cosv$: high and flat from the trigger across the window, falling
only at the ceiling. For the operator the behavior and capability curves coincide, because its detector
already demands a coherent correct product; for the router they separate at the ceiling, where the dose still
routes the language while the text collapses, so capability falls below behavior, ``leverage is not reach''
made quantitative on a benign behavior. We scope this to these directly inspectable exemplars and do not
claim global capability preservation.

\paragraph{Super weight (scale-setter, not a controller).} \emph{Type:} scale-setter.
\emph{Model:} cross-architecture phenomenon of \citet{yu2024superweight}. The globally large weight dominates
the residual norm at the beginning-of-sequence position but contributes little at generation positions, and
ablating it collapses the model without selecting any behavior. It sets the coherence budget $\Bcoh$ rather
than occupying a window, which is why it is the canonical high-leverage, no-reach object.
\emph{Example.} Ablating the weight turns a coherent answer into degenerate, repetitive text on every prompt,
with no specific behavior selected.

\paragraph{Closed window, geometric (architecture-level robustness).} \emph{Type:} closed window.
\emph{Neuron:} \neuron{25}{1790}. \emph{Model:} Gemma-3-4B. Predicted closed before dosing and confirmed:
coherence collapses at $|\cosv|<0.1$, far below any safety trigger, so no coherent dose can cross the basin.
Its near-rank-one residual (participation ratio $\approx1.1$) gives the lowest collapse ceiling we measured,
making the closed window a property of the architecture rather than of the neuron.
\emph{Example.} On a refusal prompt, every coherent dose leaves the refusal in place and the output
degenerates into repetition past the low ceiling, so no dose yields a coherent bypass.

\paragraph{Closed window, decoupled (extreme depth).} \emph{Type:} closed window (decoupled).
\emph{Neuron:} \neuron{34}{2637}. \emph{Model:} Qwen3-4B. Predicted open from its ceiling but observed
closed: at relative depth $0.94$ too few layers remain after the write for any coherent dose to propagate into
the behavior, so there is effectively no trigger. It is one of the law's informative edges, distinguishing a
ceiling-below-trigger close from a no-reachable-trigger close.
\emph{Example.} Dosing produces no behavioral change at any coherent dose and then collapses, the signature of
a write with too few downstream layers to act on the behavior.

\paragraph{Recovered dosing artifact.} \emph{Type:} sparse controller, recovered.
\emph{Neuron:} \neuron{31}{3309}. \emph{Model:} Llama-3.1-8B. Under fixed-magnitude pinning it showed the
signature of mode collapse, every prompt nominally bypassed but none coherently, so it was committed as an
artifact. The calibrated dose sweep revealed a clean window on $-\vv$; the apparent artifact was a dosing
artifact, and recovering it cost a committed verdict precisely because the control is real.
\emph{Example.} The fixed-magnitude pin returns degenerate text, read as ``destructive''; the calibrated
$-\vv$ dose returns a coherent non-refusal (harmful content withheld).

\paragraph{Prompt-affordance amplification (between a gate and no effect).} \emph{Type:} prompt-affordance
amplification. \emph{Behavior:} sort-direction (ascending instruction, target descending). \emph{Models:}
Qwen and Llama families. A matched short-rollout scan found no sort gate; the only weak signal ($0.17$) traced
to the two of twelve lists whose first element was the list maximum. An anchor-controlled follow-up showed
descending appears under perturbation almost only on those max-first lists and across many unrelated neurons,
so the effect is prompt-conditioned and neuron-promiscuous, not a sparse gate.
\emph{Example.} Prompt: ``sort in ascending order: 9, 5, 1, 7, 3'' $\to$ ``9, 7, 5, 3, 1'' (descending), but
only when the list begins with its maximum; non-max-first lists are left unsorted or scrambled.

\paragraph{No single-neuron gate found.} \emph{Type:} distributed / no gate. \emph{Behavior:} prose-to-JSON
format routing. \emph{Model:} Llama-3.1-8B. A full-layer first-token probe over $57{,}344$ neurons and a
random scan of $1{,}200$ both found no neuron producing valid JSON on a majority of prompts, while a positive
control (prompting for JSON) passed. The negative is informative because the probe is matched to the
behavior's token-local onset, so it shows such gates are uncommon under a strict criterion rather than that
the probe missed them.
\emph{Example.} A prompt asking for a JSON answer makes the best candidate emit an opening brace but not
parseable JSON on three of four prompts, while the positive control (explicitly asking for JSON) returns
valid JSON.

\paragraph{Content-risk gate (one safety example).} \emph{Type:} content-risk gate.
\emph{Neuron:} \neuron{11}{4258} (the published refusal neuron of \citet{kazemi2026}). \emph{Model:} Llama-3.1-8B.
\citet{kazemi2026} pin it to bypass safety; the constant pin also degrades capability, which the window
framework attributes to overshooting the collapse ceiling, and budget-calibrated dosing keeps the bypass
coherent (the dose-calibration correction of Section~\ref{sec:unify}). We include it only as the content-risk
instance of the taxonomy; the row-level content audit is in Appendix~\ref{app:content}.
\emph{Example.} A disallowed request (withheld): the fixed-magnitude pin bypasses refusal but degrades the
output, while the budget-calibrated dose bypasses coherently. We do not reproduce the content.

%% file: sections/appendix_trigger.tex
\section{Why a closed-form curvature trigger is not yet a law}
\label{app:trigger}

The ceiling is predicted from geometry (Section~\ref{sec:prospective}); the trigger is the under-constrained
edge. This appendix records an attempt to predict the trigger from local curvature, the corrections it
required, and the two-regime conclusion it forces. We keep it because it shows \emph{why} an attractive
closed-form theory fails and \emph{which} measurement correction matters; the main text (Section~\ref{sec:mechanism})
uses only the qualitative basin-selection mechanism and treats the trigger value as measured.

\paragraph{The candidate model.} Expanding the behavior margin along the write,
$M(\cosv)=M_0+\alpha\,\cosv+\tfrac12\beta\,\cosv^2+\cdots$ with $\beta=\vv^\top H\vv$, a silent gate
($\alpha\approx0$) flips when the quadratic term reaches the basin gap $\Delta M$, suggesting
\begin{equation}
\label{eq:trigtheory}
\cstrig \approx \sqrt{2\,\Delta M/\beta} = \tfrac{1}{\Bcoh}\sqrt{2\,\Delta M/\beta_{\mathrm{raw}}},
\qquad \beta=\beta_{\mathrm{raw}}\Bcoh^2 .
\end{equation}

\paragraph{1. Sign-averaged fits over-predict.} Fit on a symmetric (both-sign) grid with the silent
approximation, Equation~\eqref{eq:trigtheory} over-predicts the measured trigger for every audited controller,
often above the geometric ceiling $\cosv=1$ (e.g.\ $\sqrt{2|M_0|/\beta}\approx1.0$--$1.4$ for several Llama
safety pivots against measured triggers $0.4$--$0.6$). Within a behavior class the predicted exponent is also
shallow: a log--log fit of $\cstrig$ against $\beta$ gives a slope of $-0.2$ to $-0.4$ rather than the $-1/2$
the silent law requires, and the implied basin gap is not constant (coefficient of variation $\approx0.4$).

\begin{figure}[t]
\centering
\includegraphics[width=0.62\linewidth]{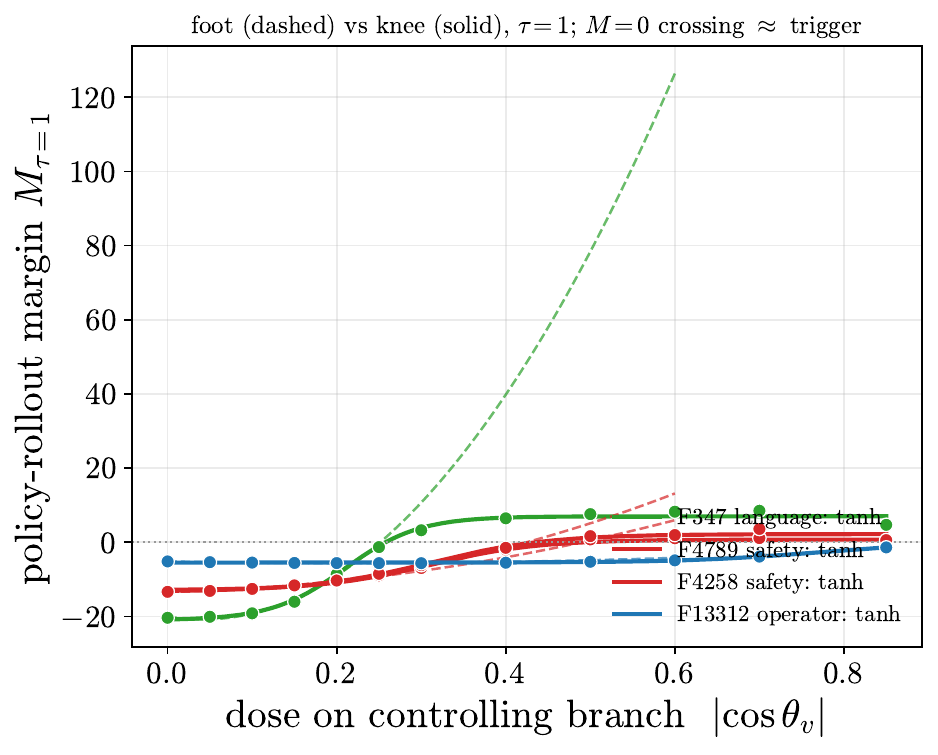}
\caption{\emph{Negative result.} Foot versus knee, branch-resolved, at $\tau{=}1$: branch resolution removes the apparent universal over-prediction but does not yield a trigger law. On the controlling branch the tanh/knee fit (solid)
crosses $M{=}0$ near the measured trigger for the language router and the refusal-bypass pivots; the origin
quadratic (dashed) reads the flat foot and overshoots. The off-axis operator (\neuron{19}{13312}) never crosses
(its margin stays below zero across the dose range). The dramatic ``always over-predicts'' of a sign-averaged
silent fit is thus largely a branch-resolution artifact.}
\label{fig:sigmoidfit}
\end{figure}

\paragraph{2. The over-prediction is largely a branch-resolution artifact.} The symmetric fit averages the
controlling and non-controlling branches and assumes $\alpha\approx0$, washing out the genuine first-order
response on the controlling sign. Re-fitting the margin on the \emph{controlling branch} alone removes most of
the over-prediction (Figure~\ref{fig:sigmoidfit}): the first-token margin already crosses near the measured
trigger for the token-local router and the refusal-bypass pivots.
\begin{center}\small
\begin{tabular}{lcc}
\toprule
neuron & measured $\cstrig$ & branch-resolved $\tau{=}1$ crossing \\
\midrule
\neuron{25}{347} (language) & $0.30$ & $0.25$--$0.26$ \\
\neuron{10}{4789} (safety/bypass) & $0.40$ & $0.49$ \\
\neuron{11}{4258} (safety/bypass) & $0.50$ & $0.43$--$0.44$ \\
\neuron{19}{13312} (operator) & no $\tau{=}1$ flip & no crossing ($>1$) \\
\bottomrule
\end{tabular}
\end{center}
This is the trigger-side echo of the sign-blind ceiling pitfall (Section~\ref{sec:contract}): trigger fits, like
ceilings, must be branch-resolved, or a sign-averaged expansion manufactures a false silent-gate over-prediction.

\paragraph{3. Foot versus knee.} The transition is sigmoidal: a quadratic fit at the origin reads only the flat
foot and overshoots, while a logistic/tanh fit captures the knee, whose $M{=}0$ crossing sits at the measured
trigger for the token-local cases and fails to cross at all for the operator (Figure~\ref{fig:sigmoidfit}).

\begin{figure}[t]
\centering
\includegraphics[width=0.62\linewidth]{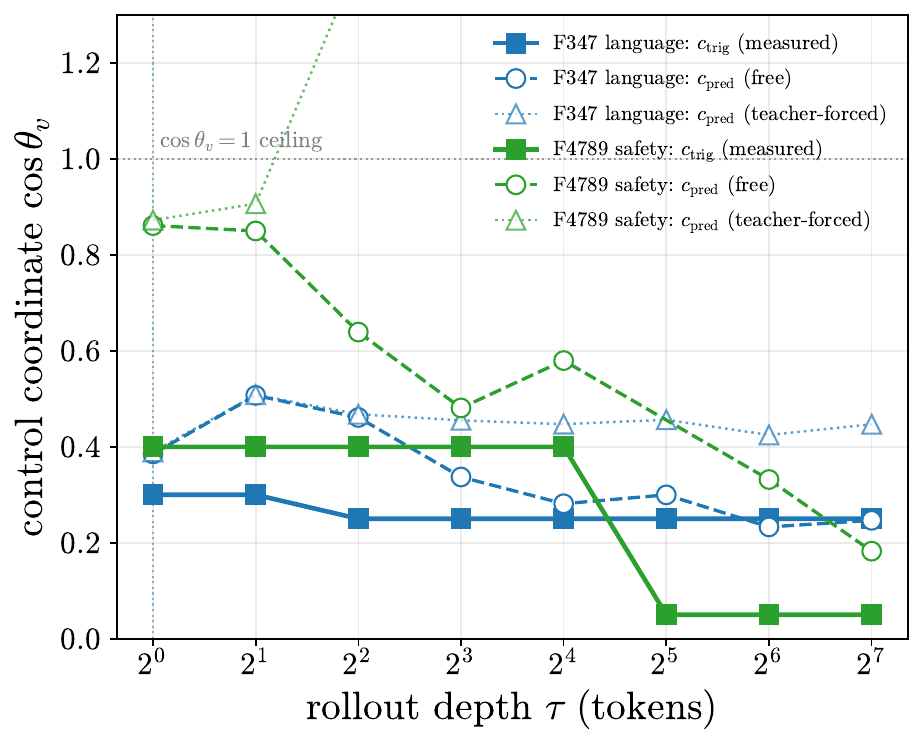}
\caption{\emph{Negative result.} Rollout-indexed trigger: rollout feedback helps token-local cases but does not give a universal trigger predictor. The first-token estimate $c_{\rm pred}(\tau{=}1)$ over-shoots; fitting the
same margin along the free rollout converges toward the measured trigger, while a teacher-forced undosed prefix
stays at the foot---so where a rollout effect exists it is autoregressive prefix feedback. Clean for the
language router; the refusal-bypass safety read is reliable only at short horizons (the trigger collapses at
$\tau\!\ge\!32$ from a non-refusal detector over-count on long generations).}
\label{fig:betahorizon}
\end{figure}

\begin{figure}[t]
\centering
\includegraphics[width=0.6\linewidth]{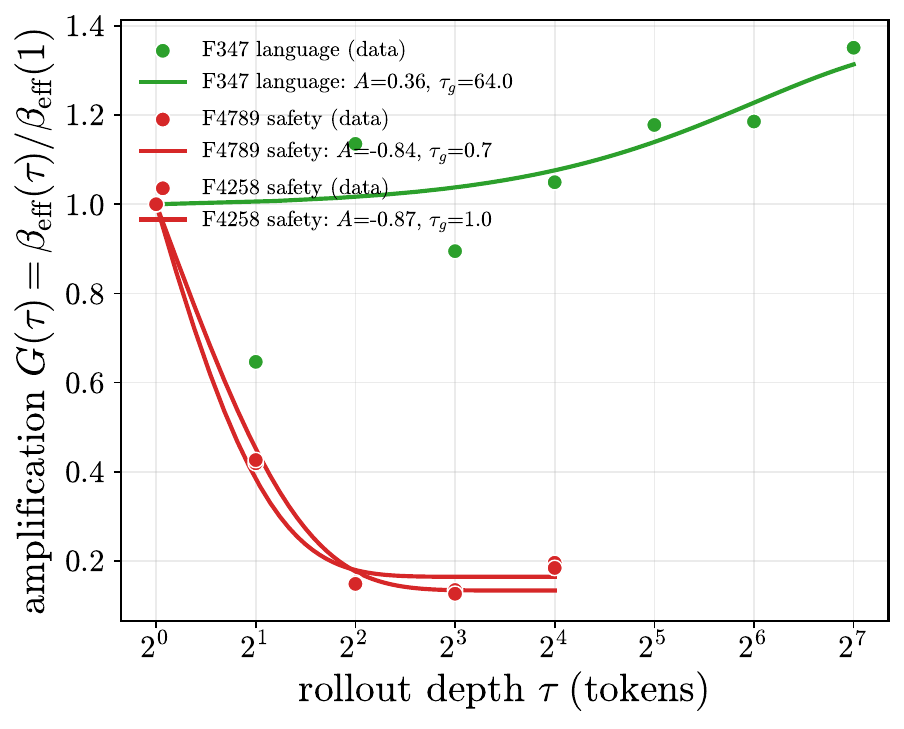}
\caption{\emph{Negative result.} Amplification is behavior-specific, so no single kernel predicts the trigger. The effective-curvature kernel $G(\tau)=\beta_{\rm eff}(\tau)/\beta_{\rm eff}(1)$
\emph{rises} for the language router ($A>0$) but \emph{attenuates} for the refusal-bypass pivots ($A<0$), so a
single amplification kernel is not universal. Exploratory; motivates the two-regime conclusion.}
\label{fig:betakernel}
\end{figure}

\paragraph{4. Rollout amplification is behavior-specific, not a universal kernel.} Reading the same margin at
later rollout horizons, the effective curvature changes with depth, but the channel differs by behavior
(Figure~\ref{fig:betahorizon}, \ref{fig:betakernel}): for the language router the controlling-branch curvature
\emph{rises} with the dosed rollout (an amplification kernel $G(\tau)=1+A(1-e^{-\tau/\tau_g})$ with $A\approx0.36$),
while a teacher-forced fixed undosed prefix stays flat---so the gain is autoregressive prefix feedback, not
static depth. For the refusal-bypass pivots the same diagnostic \emph{attenuates} ($A<0$), reflecting that the
bypass margin's well depth drifts toward zero over the rollout and that the comply-vs-refuse margin is not the
strict-actionable readout. There is thus no universal rollout-gain law.

\paragraph{5. The operator: a sequence-level confound, not a clean closed form.} The operator's product-vs-sum
margin at token position~1 is read \emph{before} the answer is emitted ($\approx\!\tau8$ in ``\,$3+4=12$''),
so its first-token failure conflates genuine rollout development with readout misalignment. Reading the
decision at the answer string, teacher-forced and rollout-free,
$M_{\mathrm{seq}}(c)=\log p(\text{product seq})-\log p(\text{sum seq})$ responds to the dose (rising from
$-3.5$ at $c{=}0$ to a peak $\approx-0.4$ on the controlling branch for \neuron{19}{13312}) but never crosses
zero, while free generation flips at $\cstrig\approx0.30$. The zero-crossing is moreover confounded by
answer length (product answers are systematically longer than sum answers, a dose-independent offset). The
operator flip is therefore part static (the dose moves the answer-level margin) and part rollout-supplied (only
free autoregressive generation completes it), with no clean closed form.

\paragraph{6. Curvature is not a discovery ranker.} Beyond the trigger predictor, curvature also fails as a
stand-alone way to \emph{find} controllers (Section~\ref{sec:gradient}). In a $200$-neuron operator-$\beta$
scan on two $36$-layer Qwen models, the highest-curvature neurons --- with $\beta_{\mathrm{raw}}$ up to sixfold
the Llama operator gate's --- produced no $K{=}1$ operator control above the noise floor, and $\beta$ was
uncorrelated with operator substitution. Ranked alone, curvature surfaces a causally inert arithmetic sector;
it explains why an off-axis controller is first-order invisible but does not itself rank controllers, the
second-order echo of the first-order salience failure of Section~\ref{sec:gradient}.

\paragraph{Conclusion: two regimes, trigger measured not predicted.} Single-neuron control has a token-local
regime, where the trigger is a branch-resolved first-token margin crossing, and a multi-token regime
(operator switching; late actionable content), where the dose moves a sequence/latent margin but free rollout
is needed to complete the flip. A universal closed-form trigger law from local curvature is not supported.
Equation~\eqref{eq:trigtheory} is retained only as a local account of gradient blindness, not a predictor.

